\documentclass[10pt,twocolumn,letterpaper]{article}

\usepackage[pagenumbers]{cvpr} %

\usepackage[dvipsnames]{xcolor}

\definecolor{cvprblue}{rgb}{0.21,0.49,0.74}
\usepackage[pagebackref,breaklinks,colorlinks,citecolor=cvprblue]{hyperref}

\def\paperID{16140} %
\def\confName{CVPR}
\def\confYear{2024}

\usepackage{multirow}
\usepackage{multicol}
\usepackage{placeins}
\usepackage{xspace}
\usepackage{xcolor}
\usepackage[accsupp]{axessibility}
\definecolor{myred}{RGB}{255,0,0}
\definecolor{mygreen}{RGB}{5,172,81}
\def\Ours{\textbf{Emu2}\xspace}
\def\OursChat{\textbf{Emu2-Chat}\xspace}
\def\OursG{\textbf{Emu2-Gen}\xspace}

\newcommand{\gr}[1]{{\textcolor{gray}{#1}}}

\newcommand{\prompt}[1]{\textcolor{black}{\small{\texttt{#1}}}}
\newcommand{\tabincell}[2]{\begin{tabular}{@{}#1@{}}#2\end{tabular}}

\makeatletter
\def\@fnsymbol#1{\ensuremath{\ifcase#1\or \dagger\or \ddagger\or
\mathsection\or \mathparagraph\or \|\or **\or \dagger\dagger
   \or \ddagger\ddagger \else\@ctrerr\fi}}
    \makeatother

\newcommand{\authorskip}{\hspace{5mm}}

\begin{document}
\title{Generative Multimodal Models are In-Context Learners}

\author{Quan Sun\textsuperscript{1}$^*$
\authorskip Yufeng Cui\textsuperscript{1}$^*$
\authorskip Xiaosong Zhang\textsuperscript{1}$^*$
\authorskip Fan Zhang\textsuperscript{1}$^*$
\authorskip Qiying Yu\textsuperscript{2,1}$^*$
\authorskip Zhengxiong Luo\textsuperscript{1} \\
\authorskip Yueze Wang\textsuperscript{1}
\authorskip Yongming Rao\textsuperscript{1}
\authorskip Jingjing Liu\textsuperscript{2}
\authorskip Tiejun Huang\textsuperscript{1,3}
\authorskip Xinlong Wang\textsuperscript{1}\thanks{Correspondence to \textit{wangxinlong@baai.ac.cn} 
}  \\[2mm]
{
\fontsize{10.4pt}{9.84pt}\selectfont
\textsuperscript{1} Beijing Academy of Artificial Intelligence \hspace{5.5mm} \textsuperscript{2} Tsinghua University \hspace{5.5mm} \textsuperscript{3} Peking University}\\[1mm]
{\fontsize{9.4pt}{9.84pt}\selectfont  \textsuperscript{$\ast$}equal contribution ~~~\textsuperscript{$\dag$}project lead} \\[2mm]
{
\fontsize{9.4pt}{9.84pt}\selectfont 
code \& models: \url{https://github.com/baaivision/Emu}
}
}

\maketitle

\begin{abstract}

The human ability to easily solve multimodal tasks in context (\textit{i.e.}, with only a few demonstrations or simple instructions), is what current multimodal systems have largely struggled to imitate. 
In this work, we demonstrate that the task-agnostic in-context learning capabilities of large multimodal models can be significantly enhanced by effective scaling-up. 
We introduce \Ours, a generative multimodal model with 37 billion parameters, trained on large-scale multimodal sequences with a unified autoregressive objective.
\Ours exhibits strong multimodal in-context learning abilities, even emerging to solve tasks that require on-the-fly reasoning, such as visual prompting and object-grounded generation.
The model sets a new record on multiple multimodal understanding tasks in few-shot settings.
When instruction-tuned to follow specific instructions, \Ours further achieves new state-of-the-art on challenging tasks such as question answering benchmarks for large multimodal models and open-ended subject-driven generation.
These achievements demonstrate that \Ours can serve as a base model and general-purpose interface for a wide range of multimodal tasks. 
Code and models are publicly available to facilitate future research.

\end{abstract}

\section{Introduction}

Multimodal tasks~\cite{li2023multimodalsurvey,gan2022zhesurvey} encompass anything involving understanding and generation in single or multiple modalities~\cite{alayrac2022flamingo,podell2023sdxl,chen2023pali3}, which can be highly diverse and long-tail.
Previous multimodal systems largely rely on designing task-specific architecture and collecting a sizable supervised training set, both of which are difficult to scale, particularly when this process needs to be repeated for each new task encountered.
By contrast, humans can solve a new task in context, \ie, with only a few demonstrations or simple instructions – a capability that current multimodal models have yet to learn. 

Recently, generative pretrained language models have demonstrated strong in-context learning abilities~\cite{brown2020language,touvron2023llama,chowdhery2022palm}. %
By training a 37-billion-parameter model \Ours and thoroughly evaluating it on diverse multimodal tasks, we demonstrate that a scaled-up multimodal generative pretrained model can harness similar in-context learning abilities and effectively generalize to unseen multimodal tasks. 
\Ours is trained with a unified autoregressive objective: predict-the-next-multimodal-element (either visual embeddings or textual tokens).
In this unified generative pretraining process, large-scale multimodal sequences (\textit{e.g.}, text, image-text pairs, and interleaved image-text-video) are used for model training.

\begin{figure*}[t]
	\centering
	\includegraphics[width=0.97\linewidth]{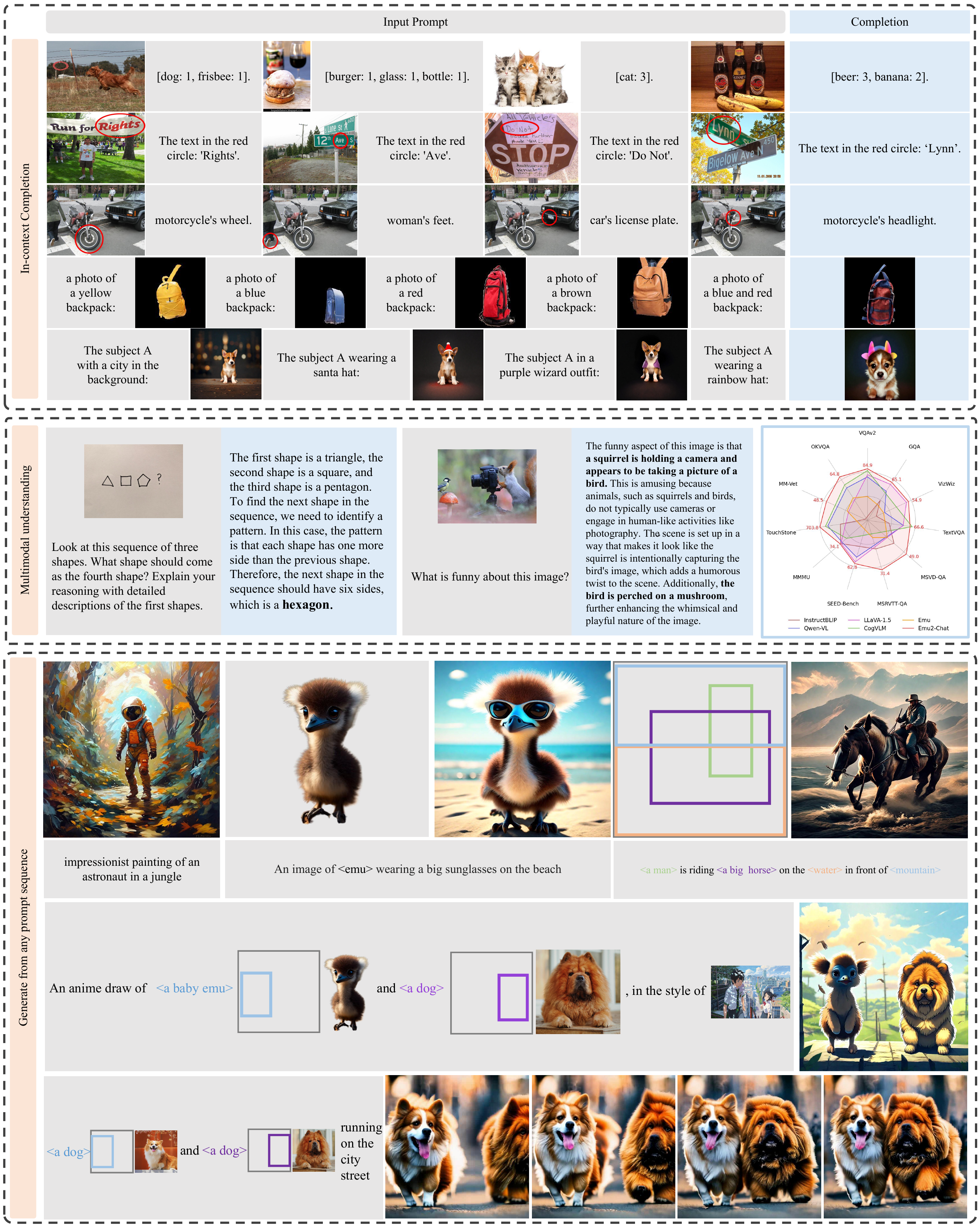}
        \vspace{-5pt}
	\caption{
 \Ours is a large generative multimodal model that serves as a foundation and a general-purpose interface for a broad range of multimodal tasks across understanding and generation, with remarkable in-context learning abilities.
 }
	\label{fig:overall_vis}
\end{figure*}

We measure \Ours's capabilities of learning from a few examples or instructions on standard multimodal datasets, as well as new tasks unseen in the training set.
Specifically,  \Ours is evaluated under two scenarios: $(a)$ \textit{few-shot setting}, where we allow as many examples as possible to fit the context window of the model;
and $(b)$ \textit{instruction tuning}, where the model is tuned to follow specific instructions.

\Ours achieves promising results in the few-shot setting on a wide range of vision-language tasks.
For example, it demonstrates state-of-the-art few-shot performance on multiple visual question-answering datasets.
We observe a performance improvement when the number of examples in context increases.
Figure~\ref{fig:overall_vis} illustrates \Ours's strong multimodal reasoning capabilities for tasks in the wild, \eg, recognition and counting in a specific format.
\Ours also learns to follow visual prompting in context (\eg, the circles laid on the images in Figure~\ref{fig:overall_vis}), even although it struggles at a smaller scale or at zero shot.

\begin{figure*}[t]
	\centering
	\includegraphics[width=0.99\linewidth]{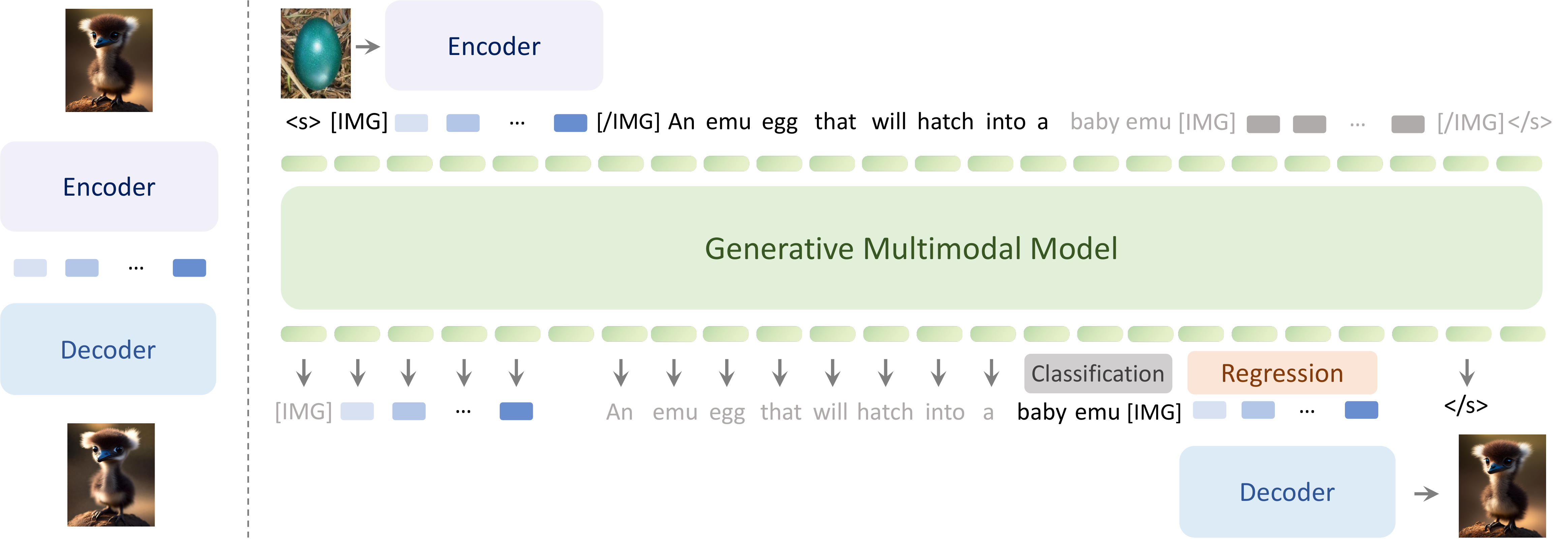}
	\caption{Overview of \Ours architecture. \Ours learns with a predict-the-next-element objective in multimodality. Each image in the multimodal sequence is tokenized into embeddings via a visual encoder, and then interleaved with text tokens for autoregressive modeling. The regressed visual embeddings will be decoded into an image or a video by a visual decoder.}
	\label{fig:method}
\end{figure*}

As \Ours is inherently equipped to handle interleaved text-image-video at both input and output, it serves as a powerful and versatile base model for diverse multimodal tasks, 
by following specific task instructions.
For example, after instruct tuning with conversational data, \Ours achieves state-of-the-art results on visual question-answering tasks, and surpasses previous models of more complex designs.
In addition, \Ours can be fine-tuned to function as a controllable visual generation model of high quality.
It is capable of accepting a mixture of text, locations and images as conditions, and generating images that are grounded as specified.

Given the broad spectrum of capabilities displayed by \Ours, we conduct a thorough analysis of its potential societal implications and discuss in detail potential concerns over misuse.
By identifying further tasks where \Ours's in-context learning can further improve,
we highlight the necessity for continuous enhancement of the model and the importance of deploying \Ours responsibly.

\section{Approach}

\subsection{Model Architecture}
\Ours is a generative multimodal model that learns with a predict-the-next-element objective in multimodal context.
As illustrated in~\ref{fig:method}, the architecture of \Ours consists of three components: Visual Encoder, Multimodal Modeling, and Visual Decoder.
Each image in the input multimodal sequence is tokenized into continuous embeddings via the Visual Encoder and then interleaved with text tokens for autoregressive Multimodal Modeling. 
The regressed visual embeddings are then decoded into an image or a video by the Visual Decoder.
Specifically, we leverage pretrained EVA-02-CLIP-E-plus~\citep{sun2023evaclip}, LLaMA-33B~\citep{touvron2023llama} and SDXL~\citep{podell2023sdxl} to initialize the Visual Encoder, Multimodal Modeling, and Visual Decoder, respectively.
Compared to Emu~\citep{sun2023generative}, \Ours embraces a simpler framework which connects the Visual Encoder and Multimodal Modeling through mean pooling each image to $8 \times 8$ image patches, followed by a linear projection, instead of using an additional C-Former~\cite{sun2023generative}.

\subsection{Pretraining}

\subsubsection{Data}

The pretraining data for \Ours comprises several publicly accessible datasets, including image-text pairs from LAION-2B~\citep{schuhmann2022laion} and CapsFusion-120M~\cite{yu2023capsfusion}, video-text pairs from WebVid-10M~\citep{bain2021frozen}, interleaved image-text data from Multimodal-C4 (MMC4)~\citep{mmc4}, interleaved video-text data from YT-Storyboard-1B~\citep{sun2023generative}, grounded image-text pairs from GRIT-20M introduced by Kosmos-2~\citep{peng2023kosmos} and CapsFusion-grounded-100M curated by CapsFusion-120M. Additionally, language-only data from Pile~\citep{pile} is included to retain textual reasoning capability.

\subsubsection{Training}

Similar to Emu~\citep{sun2023generative}, \Ours learns with the predict-the-next-element objective within a multimodal sequence. Each image is encoded into $N=64$ dimension-fixed visual embeddings and then interleaved with text tokens to construct a multimodal sequence. The interleaved sequence is then fed into a Transformer decoder for autoregressive modeling.

\Ours is first pretrained on image-text and video-text pair data with only captioning loss on the text tokens. 
The input images are resized to $224 \times 224$. We adopt the AdamW optimizer~\citep{loshchilov2017decoupled} with \(\beta_1 = 0.9\), \(\beta_2 = 0.95\), \(\epsilon = 1 \times 10^{-6}\). The maximum learning rate is \(1 \times 10^{-4}\) for the linear projection layer, \(3 \times 10^{-5}\) for Multimodel Modeling, and \(5 \times 10^{-5}\) for Visual Encoder. We pretrain \Ours on 162 million image-text samples and 7 million video-text samples for 35,200 iterations. The global batch size is 6,144 for the image-text pairs and 768 for video-text pairs. The training process is then restarted at a higher 448-pixel resolution for an additional 4,000 iterations.

Then, we freeze the Visual Encoder and only optimize the linear projection layer and Multimodel Modeling with both text classification loss and image regression loss. 
Additional datasets including image-text interleaved data, video-text interleaved data, grounded image-text pair data, and language-only data are used in the training. 
All images are resized to $448 \times 448$, and the maximum learning rate is \(1 \times 10^{-5}\). We use a global batch size of 12,800 for image-text pair data, 6,400 for video-text pair data, 3,200 for image-text and video-text interleaved data, and 800 for language-only data. The training process spans 20,350 iterations and consumes about 160 million samples of image-text data and 3.8B tokens of language-only data.

\subsubsection{Visual Decoding} 
We train the Visual Decoder to directly decode visual embeddings generated by the Visual Encoder into image. 
We use SDXL-base\citep{podell2023sdxl} as the initialization of our Visual Decoder, which is fully trained to solve the new task of autoencoding. 
Specifically, we use $N$ visual embeddings as the condition input to the Visual Decoder and adjust the dimension of the projection layers in cross-attention modules to match the dimension of visual embeddings.

Unlike Emu~\citep{sun2023generative} where each optimization step of its Visual Decoder requires an autoregressive inference of the language model,
\Ours's visual decoding can be considered as training a detokenizer, which can be trained off-the-shelf without the language model.
Once trained, the Visual Decoder together with the Visual Encoder works as an image autoencoder that can tokenize an image into embeddings and detokenize back.
During \Ours inference, it generates $N$ image embeddings and decodes to an image on the fly.

For the decoding of video data, we train a diffusion-based decoder~\citep{make-a-video}. Similar to~\citep{modelscope,videogen}, we adapt a 2D denoising U-Net to 3D style by inserting a 1D temporal convolution following each 2D spatial convolutional layer and extending the spatial attention to spatial-temporal attention. This video decoder is initialized via Stable Diffusion 2.1~\citep{rombach2022high} and fully trained to generate video clips conditioned on visual embeddings from \Ours.

\noindent\textbf{Training Setup.}
We use the images in LAION-COCO~\cite{laioncoco} and LAION-Aesthetics~\citep{laionaesthetics} to train the Visual Decoder under the task of image autoencoding. 
The Visual Encoder and VAE in SDXL are frozen, and only the U-Net is updated during training. 
We adopt AdamW optimizer~\citep{loshchilov2017decoupled} with $\beta_1=0.9,\beta_2=0.999$ and the weight decay of 0.01. 
We use $log$ learning rate warm-up and linear learning rate decay with a peak learning rate of \(1 \times 10^{-4}\) for 2,000 and 6,000 steps, respectively. 
We filter out images whose resolution is lower than $512 \times 512$. The input to the Visual Encoder is set to $448 \times 448$, while the output of the Visual Decoder is set to $1024 \times 1024$. 
We also employ the classifier-free guidance~\citep{ho2022classifier}, which randomly discards image embeddings with the probability of $10\%$. 
The batch size is set to 2,048 in total.

\subsection{Instruction Tuning}

\Ours can be efficiently aligned to follow specific task instructions.
We fine-tune the base model with conversational data to yield \OursChat, which is capable of following multimodal questions and making responses in dialogue.
Similarly, we derive a controllable visual generation model \OursG, which is capable of accepting a mix of text, locations, and images as conditions, and generating images that are grounded in the specified text or subject.

\begin{figure}[t]
	\centering
	\includegraphics[width=0.95\linewidth]{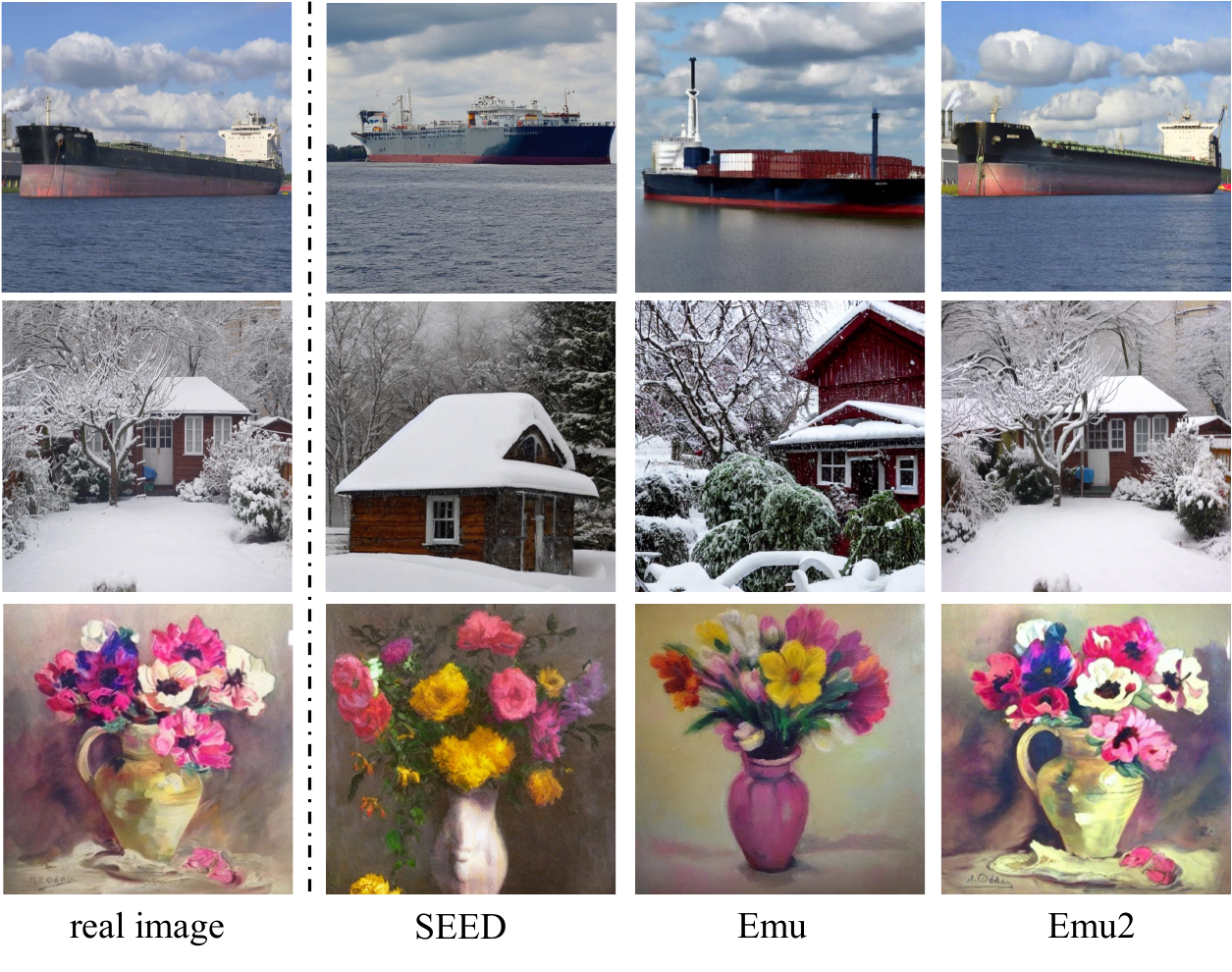}
	\caption{Comparison of autoencoding results among different methods~\citep{ge2023planting,sun2023generative}. 
            \Ours's Visual Encoder and Visual Decoder in the architecture of CLIP-Diffusers form a strong autoencoder.}
	\label{fig:vis_aotoencoding}
\end{figure}

\subsubsection{Instruction-Following Chat}

\noindent\textbf{Training Data.}
We adopt a uniform approach to train on both academic-task-oriented datasets and multimodal chat data to empower \OursChat with the instruction-following ability while retaining rich visual knowledge. 
As academic-task-oriented datasets have brief annotations that limit the model's capacity to provide more comprehensive and helpful responses, we distinguish between these two data categories by employing different system messages and including instructions with output-format control information as used in \cite{liu2023improved}.
A summary of data used is as follows:
$(a)$ Academic-task-oriented data: image captioning~\citep{chen2015microsoft,sidorov2020textcaps}, visual question answering~\citep{goyal2017making,hudson2019gqa,singh2019towards}, knowledgeable question answering~\citep{marino2019ok,lu2022learn}, multimodal classification~\citep{li2023m}, and referring expression comprehension~\citep{kazemzadeh2014referitgame,mao2016generation}.
$(b)$ Multimodal chat data: GPT-assisted visual instruction~\citep{liu2023visual,zhang2023llavar}, language instruction~\citep{sharegpt,taori2023stanford}, clock reading~\citep{yang2022s}, and video chat~\citep{li2023videochat}.

\noindent\textbf{Training Objective.}
In instruction tuning of \OursChat, two special tokens, \texttt{[USER]} and \texttt{[ASSISTANT]}, are incorporated into the model to denote roles. These tokens help organize different data types in the following format: ``{\prompt{<Sys.Msg.> [USER]: <Instruction> [ASSISTANT]: <Answer>}}''.
Here \prompt{\text{<Sys.Msg.>}} represents system message and varies between the two major task categories (academic-task-oriented and multimodal chat). The \prompt{\text{<Instruction>}} section comprises multimodal tokens, including images, videos, and text. Only tokens in the \prompt{\text{<Answer>}} section will be supervised by cross-entropy loss during training. 

\noindent\textbf{Training Setup.}
We use a global batch size of 768 and train for 8k steps. The learning rate linearly warms up to \(1 \times 10^{-5}\) in the first 100 steps, then decays to zero with a cosine schedule. The model is trained using the AdamW optimizer with \(\beta_1 = 0.9\), \(\beta_2 = 0.98\), \(\epsilon = 1 \times 10^{-6}\), and a gradient clipping of 5.0. The sequence length during training is limited to 2048, and any excess beyond that is truncated directly. We consistently employed an input image/video resolution of 448 $\times$ 448. For video data, we uniformly sample frames in time as input to the model. The number of sampled frames for each video is randomly chosen from 8, 12, and 16. To capture more intricate spatial details, following the visual encoder stage, we apply mean-pooling to each static image, dividing it into 16 $\times$ 16 tokens during instruction fine-tuning. This differs from the pre-training phase, where 8 $\times$ 8 tokens were utilized.

\subsubsection{Controllable Visual Generation}

\noindent\textbf{Training Data.}  
We leverage a mix of high-quality datasets to unleash the potential of controllable generation in context.
We use a grounded image-text pair dataset CapsFusion-grounded-100M and GRIT~\citep{peng2023kosmos} for grounded text-to-image generation. 
To mitigate the impact of image backgrounds on the effectiveness of multi-entity subject-driven generation, we employ SAM~\citep{kirillov2023segment} to preprocess the grounding data, yielding a subset of approximately 5 million samples with segmentation results. 
Additionally, we leverage InstructPix2Pix constructed by \citep{brooks2023instructpix2pix} for image editing tasks. 
For the text-to-image task, we use a filtered subset of CapsFusion~\citep{yu2023capsfusion}, LAION-Aesthetics~\citep{laionaesthetics}, SA-1B~\citep{kirillov2023segment}, and LAION-High-Resolution~\citep{laionhighresolution}. 

We also collect data from premium sources (\eg, Unsplash~\citep{unsplash}) and outputs from advanced text-to-image systems (\eg, Midjourney-V5~\citep{Midjourney} and DALL-E-3~\citep{betker2023dalle3}) for quality fine-tuning. 
This diverse dataset includes around 500k high-quality image-text pairs.
For all the data above, during the training, only samples with image resolutions higher than $448 \times 448$ were retained to ensure generation quality. More details can be found in the supplementary.

\begin{table}[htbp]
\centering
\resizebox{1.0\linewidth}{!}{
\setlength{\tabcolsep}{0.1cm}
\begin{tabular}{l|cccccc}
\toprule 
Model 
& Shot  & VQAv2   & OKVQA   & VizWiz   & TextVQA  & \tabincell{c}{Hateful\\Memes} \\
\midrule
\multirow{3}{*}{ Kosmos-1 (1.6B) } 
& 0  & 51.0 & - & 29.2 & - & - \\
& 4  & 51.8 & - & 35.3 & - & - \\
& 8  & 51.4 & - & 39.0 & - & - \\ 
\midrule
\multirow{5}{*}{ Flamingo (9B) } 
& ~~0$^*$  & 51.8 & 44.7 & 28.8 & 31.8 & 57.0 \\ 
& 4  & 56.3 & 49.3 & 34.9 & 33.6 & 62.7 \\ 
& 8  & 58.0 & 50.0 & 39.4 & 33.6 & 63.9 \\ 
& 16 & 59.4 & 50.8 & 43.0 & 33.5 & 64.5 \\ 
\midrule
\multirow{4}{*}{ Flamingo (80B) } 
& ~~0$^*$  & 56.3 & 50.6 & 31.6 & 35.0 & 46.4 \\
& 4  & 63.1 & 57.4 & 39.6 & 36.5 & \underline{68.6} \\
& 8  & 65.6 & \underline{57.5} & 44.8 & 37.3 & \textbf{70.0} \\ 
& 16 & 66.8 & \bf{57.8} & 48.4 & 37.6 & \textbf{70.0} \\ 
\midrule
\multirow{4}{*}{ IDEFICS (80B) } 
& ~~0$^*$  & 60.0 & 45.2 & 36.0 & 30.9 & 60.6 \\
& 4  & 63.6 & 52.4 & 40.4 & 34.4 & 57.8 \\
& 8  & 64.8 & 55.1 & 46.1 & 35.7 & 58.2 \\
& 16 & 65.4 & 56.8 & 48.3 & 36.3 & 57.8 \\
\midrule
\multirow{4}{*}{~~Emu~~(14B) }
& ~~0$^*$  & 52.9 & 42.8 & 34.4 & -    & -    \\
& 4  & 58.4 & -    & 41.3 & -    & -    \\
& 8  & 59.0 & -    & 43.9 & -    & -    \\
& 16 & -    & -    & -    & -    & -    \\
\midrule
\multirow{4}{*}{~~\Ours~~\textbf{(37B)} }
& 0  &  33.5 & 26.7 & 40.4 & 26.4 & 52.2 \\
& 4  &  67.0 & 53.2 & 54.6 & 48.2 & 62.4 \\
& 8  &  \underline{67.8} & 54.1 & \underline{54.7} & \underline{49.3} & 65.8 \\
& 16 &  \textbf{68.8} & 57.1 & \textbf{57.0} & \textbf{50.3} & 66.0 \\
\bottomrule
\end{tabular}
}
\caption{Zero-shot and few-shot evaluations of \Ours. $0^*$ denotes text two-shot and image zero-shot results following Flamingo~\cite{alayrac2022flamingo}. The best results are in \textbf{bold} and the second best are \underline{underlined}.}
\label{tab:fewshot}
\end{table}

\begin{table*}[htbp]
\centering
\small
\setlength{\tabcolsep}{0.15cm}
\resizebox{1.0\linewidth}{!}{
\begin{tabular}{l|ccccc|cc|cccc}
\toprule
\multirow{2}{*}{Model} 
& \multicolumn{7}{|c}{Visual Question Answer} & \multicolumn{4}{|c}{LMM Benchmarks} \\
& \tabincell{c}{VQAv2\\\citep{goyal2017making}} & \tabincell{c}{OKVQA\\\citep{marino2019ok}} & \tabincell{c}{GQA\\\citep{hudson2019gqa}}  & \tabincell{c}{VizWiz\\\citep{gurari2018vizwiz}}  & \tabincell{c}{TextVQA\\\citep{singh2019towards}} & \tabincell{c}{MSVD\\\citep{xu2017video}}  & \tabincell{c}{MSRVTT\\\citep{xu2017video}} & \tabincell{c}{SEED\\\citep{li2023seed}}   & \tabincell{c}{MM-Vet\\\citep{yu2023mm}} & \tabincell{c}{TS\\\citep{bai2023touchstone}} & \tabincell{c}{MMMU\\\citep{yue2023mmmu}} \\
\midrule
Flamingo-9B~\citep{alayrac2022flamingo}
& 51.8 & 44.7 & -    & 28.8 & -    & 30.2 & 13.7     
& -    & -    & -    & -    \\
Flamingo-80B~\citep{alayrac2022flamingo} 
& 56.3 & 50.6 & -    & 31.6 & -    & 35.6 & 17.4      
& -    & -    & -    & -    \\
Kosmos-1~\citep{huang2023language} 
& 51.0 & -    & -    & 29.2 & -    & -    & -      
& -    & -    & -    & -    \\
Kosmos-2~\citep{peng2023kosmos}
& 51.1 & -    & -    & -    & -    & -    & -      
& 50.0 & -    & -    & 26.6 \\
BLIP-2-13B~\citep{li2023blip}
& -    & -    & 41.0 & 19.6 & 42.5 & 20.3 & 10.3
& 46.4 & 22.4 & -    & -    \\
InstructBLIP-13B~\citep{dai2023InstructBLIP}
& -    & -    & 49.5 & 33.4 & 50.7 & 41.2 & 24.8
& -    & 25.6 & 552.4& -    \\
IDEFICS-9B~\citep{laurenccon2023obelics}
& 50.9 & 38.4 & -    & 35.5 & 25.9 & -    & - 
& -    & -    & -    & -    \\
IDEFICS-80B~\citep{laurenccon2023obelics}
& 60.0 & 45.2 & -    & 36.0 & 30.9 & -    & - 
& -    & -    & -    & -    \\
Shikra-13B~\citep{chen2023shikra}
&~~77.4*& 47.2 & -   & -    & -    & -    & - 
& -    & -    & -    & -    \\
Qwen-VL-13B-Chat~\citep{bai2023qwen} 
&~~78.2*&~~56.6*&~~57.5* & 38.9 &~~61.5*& -    & - 
& 58.2 & -    & 645.2& -    \\
LLaVA-1.5-13B~\citep{liu2023improved}
&~~80.0*& -    &~~63.3*& 53.6 & 61.3 & -    & - 
& 61.6 & 35.4 & -    & 33.6 \\
CogVLM~\citep{wang2023cogvlm}
& ~~83.4* & ~~58.9* & -     & -    & ~~\bf{68.1}*  & -    & -    
& -   & -    & 662.6& 30.1 \\
Emu-I~\citep{sun2023generative}
& 62.0 & 49.2 & 46.0 & 38.3 & -    & 37.0 & 21.2
& -    & 36.3 & -    & -    \\
\OursChat  
&~~\bf{84.9}*&~~\bf{64.8}*&~~\bf{65.1}*& \bf{54.9} &~~66.6*& \bf{49.0} & \bf{31.4} 
& \bf{62.8} & \bf{48.5} & \bf{703.8} & \bf{34.1} \\
\bottomrule
\end{tabular}
}
\caption{Results on visual question answering and LMM benchmarks. * indicates that samples from this task's training set have been trained. SEED and TS respectively represent SEED-Bench~\cite{li2023seed} and TouchStone~\cite{bai2023touchstone}. For MM-Vet, we present the average result of five scoring runs.}
\label{tab:chatperformance}
\end{table*}

\noindent\textbf{Training Objective.}
We use the same unified generative pretraining objective to adapt to diverse generation tasks in context.
Specifically, a training sample for generation is formulated as: 
``{\prompt{<s>A photo of <p>a man</p><coor>image embedding of object localization image</coor>[IMG]image embedding of man[/IMG]sitting next to <p>a dog</p><coor>image embedding of object localization image</coor>[IMG]image embedding of dog[/IMG][IMG]image embedding of the whole image[/IMG]</s>}}''. 
We represent the coordinates of each object directly in image form by drawing the bounding box of each object at its specified location on a black image. Our \OursG conducts unified multimodal modeling of the text, object image, and corresponding object localization image.
The regression loss only applies to the visual embeddings of the last image. 
We freeze the Visual Encoder during fine-tuning.
We randomly drop tokens of entities and object localization image to enhance model adaptability and robustness.
Additionally, we apply data augmentation to each object image, incorporating random background variations and random crop, aiming to reduce the reliance on image backgrounds.

\noindent\textbf{Training Setup.}  
We use a global batch size of 4,096 and train for 3k steps. The learning rate linearly warms up to \(5 \times 10^{-5}\) in the first 100 steps, then decays to zero with a cosine schedule. 
We further fine-tune for 900 steps using the 500k high-quality pairs with a batch size of 2048.

\section{Evaluation}

\subsection{Pretrained Base Model}
We evaluate zero-shot and few-shot abilities of \Ours on OKVQA~\citep{marino2019ok}, VQAv2~\citep{goyal2017making}, VizWiz~\citep{gurari2018vizwiz}, TextVQA~\citep{singh2019towards}, and HatefulMemes~\citep{kiela2020hateful} tasks. Details of the datasets and prompts can be found in supplementary materials. The results are presented in Table~\ref{tab:fewshot}. \Ours demonstrates remarkable in-context ability, showcasing improved performance with more in-context samples seen.
Specifically, on VQAv2, VizWiz and TextVQA datasets, \Ours outperforms Flamingo-80B and IDEFICS-80B under all few-shot settings with a much smaller model scale (37B).

Figure~\ref{fig:overall_vis} demonstrates \Ours's few-shot capabilities in the wild.
For example, the model learns to classify and count simultaneously in a specific format via a few examples (row 1). Additionally, \Ours is capable of following visual prompts in context, \eg, the red circles laid on the images (row 2 and 3).

\subsection{Instruction-Following Chat}

Our \OursChat is evaluated on academic-task-oriented benchmarks including image question-answering datasets (VQAv2~\citep{goyal2017making}, OKVQA~\citep{marino2019ok}, GQA~\citep{hudson2019gqa}, VizWiz~\citep{gurari2018vizwiz}, TextVQA~\citep{singh2019towards}) and video question-answering datasets (MSVD~\citep{xu2017video} and MSRVTT~\citep{xu2017video}). The evaluation also encompassed recent benchmarks for large multimodal models, including SEED-Bench~\citep{li2023seed}, MM-Vet~\citep{yu2023mm}, TouchStone~\citep{bai2023touchstone} and MMMU~\citep{yue2023mmmu}. When evaluated on SEED-Bench, we followed the setup of LLaVa-1.5~\citep{liu2023improved} by presenting options to the model for completing multiple-choice tasks.

As shown in Table~\ref{tab:chatperformance}, \OursChat consistently outperforms other models in image question-answering tasks, encompassing well-established benchmarks like VQAv2 and GQA. Notably, it shows a noticeable improvement in the OKVQA task, which requires the utilization of external knowledge, showcasing the advantage of our model for mastering real-world knowledge. 
For video question-answering, \OursChat demonstrated advantages even though it did not use video question-answering data for training. It achieved an accuracy of 49.0 and 31.4 on the MSVD-QA and MSRVTT-QA tasks, respectively, surpassing InstructBLIP and the larger Flamingo-80B. More importantly, our model has also achieved better results on LMM benchmarks. LMM benchmarks such as MM-Vet provide a more comprehensive evaluation of model abilities, including solving complicated tasks. \OursChat achieves a score of 48.5 in MM-Vet and 703.8 in TouchStone, confirming its superior capability in understanding and solving multimodal problems. 

In addition, we demonstrated the visual grounding capability of our model using the refer expression comprehension benchmarks. In Table~\ref{tab:chatrec}, \OursChat achieved the best results among generalist models on RefCOCO~\citep{kazemzadeh2014referitgame}, RefCOCO+~\citep{mao2016generation} and RefCOCOg~\citep{mao2016generation}. 
Its most notable advantage was observed in RefCOCO+, which focused solely on purely appearance-based descriptions without allowing the use of position references. This highlights our model's powerful perceptual abilities in capturing intricate details.

\begin{table}[t]
\centering
\small
\setlength{\tabcolsep}{0.10cm}
\resizebox{\linewidth}{!}{
\begin{tabular}{l|cccccccc}
\toprule
\multirow{2}{*}{Model} & \multicolumn{3}{|c}{RefCOCO} & \multicolumn{3}{c}{RefCOCO+} & \multicolumn{2}{c}{RefCOCOg} \\
& val & testA & testB & val & testA & testB & val & test \\
\midrule
OFA-L~\citep{wang2022ofa}
& 79.96 & 83.67 & 76.39 
& 68.29 & 76.00 & 61.75 
& 67.57 & 67.58 \\
Shikra-7B~\citep{chen2023shikra} 
& 87.01 & 90.61 & 80.24 
& 81.60 & 87.36 & 72.12 
& 82.27 & 82.19 \\
Shikra-13B~\citep{chen2023shikra} 
& 87.83 & 91.11 & 81.81 
& 82.89 & 87.79 & 74.41 
& 82.64 & 83.16 \\
Qwen-VL-7B~\citep{bai2023qwen}
& 89.36 & 92.26 & 85.34
& 83.12 & 88.25 & 77.21
& 85.58 & 85.48 \\
\OursChat    
& \bf{90.40} & \bf{93.88} & \bf{85.97}
& \bf{87.05} & \bf{91.43} & \bf{80.47}
& \bf{87.64} & \bf{88.11} \\
\midrule
\gr{CogVLM~\citep{wang2023cogvlm}}
& \gr{92.51} & \gr{93.95} & \gr{88.73}
& \gr{87.52} & \gr{91.81} & \gr{81.43}
& \gr{89.46} & \gr{90.09} \\
\bottomrule
\end{tabular}
}
\caption{Results on referring expression comprehension. We grayed out CogVLM because its generalist grounding-enhanced model was specialist trained on high-quality grounding data.}
\label{tab:chatrec}
\vspace{-10pt}
\end{table}

\begin{figure*}[t]
	\centering
	\includegraphics[width=0.999\linewidth]{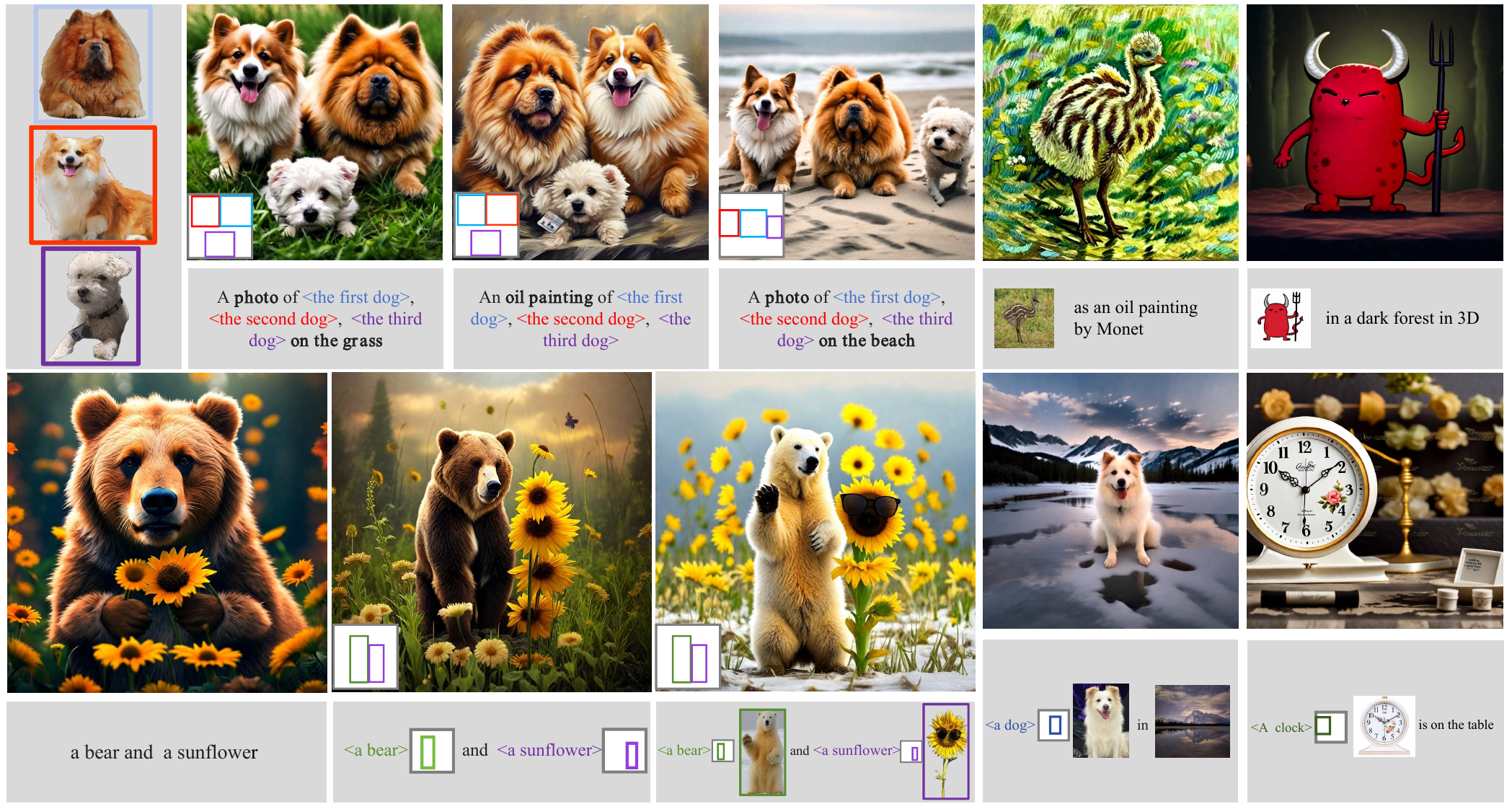}
	\caption{Visualization of \OursG's controllable generation capability. The model is capable of accepting a mix of text, locations and images as input, and generating images in context.
    The presented examples include text- and subject-grounded generation, stylization, multi-entity composition, subject-driven editing, and text-to-image generation.}
\label{fig:overall_vis_more}
\end{figure*}

\subsection{Controllable Visual Generation}

\noindent\textbf{Qualitative Results.} 
Figure~\ref{fig:vis_aotoencoding} presents a visualization of \Ours's autoencoding results. 
With \Ours's Visual Encoder and Visual Decoder, we can tokenize an image into visual embeddings and detokenize them back.
Compared with SEED~\citep{ge2023planting} and Emu~\citep{sun2023generative},  \Ours shows significantly superior results.
We also evaluate our image autoencoding results on MS-COCO~\cite{lin2014microsoft} and achieve a strong 0.907 CLIP-I~\citep{radford2021clip} score.
More results are in the supplementary.

As depicted in Figure~\ref{fig:overall_vis_more},  \OursG is capable of accepting a mixture of text, locations and images as input, and generating images in context.
The model skillfully engages in various controllable visual generation tasks in a zero-shot setting, capitalizing on the in-context learning capabilities  in multimodality. 
Examples in Figure~\ref{fig:overall_vis_more} show generated images of three dogs conditioned on different subjects, locations and scenarios.
The presented visual samples demonstrate the model's proficiency in tasks such as re-contextualization, stylization, modification, region-controllable generation, and multi-entity composition.

\noindent\textbf{Zero-shot Text-to-image Generation.} 
We evaluate the zero-shot text-to-image generation capability on 30k randomly sampled data from the MS-COCO~\citep{lin2014microsoft} validation set. We employ CLIP-ViT-B~\citep{radford2021learning}, following the approach in DALL-E 3\cite{betker2023dalle3}, to calculate the CLIP-T score to assess prompt-following ability. 
Additionally, we utilize CLIP-ViT-L, as in GILL\cite{koh2023generating}, to compute the CLIP-I score for measuring image similarity. 
A higher score means the generated image is more similar to the prompt or the real image.
Table~\ref{tab:text2image_fid} shows that \OursG achieves the state-of-the-art performance in terms of both CLIP-I and CLIP-T scores compared to various unimodal generation models and multimodal models.
More text-to-image generation cases can be found in supplementary.

\begin{table}[t]
    \centering
    \small
    \resizebox{0.66\linewidth}{!}{
        \begin{tabular}{l|c|c}
            \toprule
            Models                                  & CLIP-I $\uparrow$ & CLIP-T $\uparrow$ \\
            \midrule
            \multicolumn{3}{c}{\textit{unimodal generation models}} \\
            \midrule
            MUSE~\citep{chang2023muse}                     & - & \bf0.320 \\
            Imagen~\cite{saharia2022photorealistic}        & - & 0.270 \\
            DALL-E 2 \dag ~\citep{ramesh2022hierarchical}  & - & 0.314 \\
            DALL-E 3 \dag ~\citep{betker2023dalle3}        & - & \bf0.320 \\ 
            SDv1.5~\citep{rombach2022high}                 & 0.667 & 0.302 \\
            SDXL~\citep{podell2023sdxl}                    & 0.674 & 0.310 \\
            \midrule
            \multicolumn{3}{c}{\textit{multimodal generation models}} \\
            \midrule
            GILL~\citep{koh2023generating}                 & 0.684 & - \\
            SEED~\citep{ge2023planting}                    & 0.682 & - \\
            Emu~\citep{sun2023generative}                  & 0.656 & 0.286 \\
            \OursG & \bf0.686 & 0.297  \\
            \bottomrule
        \end{tabular}
    }
    \caption{Quantitative comparison of zero-shot text-to-image generation on MS-COCO~\citep{lin2014microsoft} validation set. 30k samples are randomly sampled. \dag CLIP-T score is calculated on 4,096 samples. We also evaluate our image autoencoding results on MS-COCO which achieves a strong 0.907 CLIP-I score.}
    \label{tab:text2image_fid}
\end{table}

\noindent\textbf{Zero-shot Subject-driven Generation.} 
Following Kosmos-G~\citep{pan2023kosmos}, we also evaluate our model's subject-driven image editing ability on DreamBench~\citep{ruiz2023dreambooth}.
We generate four images for each prompt, resulting in a total of 3,000 images for a comprehensive evaluation. 
We employ DINO~\citep{caron2021emerging} and CLIP-I~\citep{radford2021clip} to evaluate subject fidelity, and CLIP-T~\citep{radford2021clip} to evaluate text fidelity, aligning with the methodology established by DreamBooth. 
Notably, \OursG excels in subject fidelity, as evidenced by its superior performance on DINO and CLIP-I metrics compared to methods like BLIP-Diffusion and Kosmos-G. 
\OursG impressively reconstructs subjects with just one image input in zero-shot setting, demonstrating superior subject fidelity through powerful visual decoding. Further illustrative cases are provided in the supplementary, showcasing \OursG's proficiency in multi-entity generation.

\begin{table}[t]
\centering
\small
\resizebox{\linewidth}{!}{
\setlength{\tabcolsep}{0.1cm}
\begin{tabular}{lccc}
\toprule
Methods & DINO $\uparrow$ & CLIP-I $\uparrow$ & CLIP-T $\uparrow$ \\
\midrule Real Images (Oracle) & 0.774 & 0.885 & - \\
\midrule \multicolumn{4}{c}{ \textit{Fine-Tuning} } \\
\midrule Textual Inversion ~\citep{gal2022image} & 0.569 & 0.780 & 0.255 \\
DreamBooth ~\citep{ruiz2023dreambooth} & 0.668 & 0.803 & \bf0.305 \\
BLIP-Diffusion ~\citep{li2023blip} & 0.670 & 0.805 & 0.302 \\
\midrule \multicolumn{4}{c}{ \textit{Test Time Tuning Free} } \\
\midrule Re-Imagen*~\citep{chen2022re} & 0.600 & 0.740 & 0.270 \\
SuTI~\citep{chen2023subject}  & 0.741 & 0.819 & 0.304 \\
BLIP-Diffusion* ~\citep{li2023blip} & 0.594 & 0.779 & 0.300 \\
Kosmos-G* (single image input) & 0.694 & 0.847 & 0.287 \\

\OursG* (single image input) & \bf0.766 & \bf0.850 & 0.287  \\
\bottomrule
\end{tabular}
}
\caption{Quantitative comparison of zero-shot single-entity subject-driven generation on DreamBench. * denotes zero-shot methods.}
\label{tab:dreambench}
\end{table}

\section{Related Work}

\noindent\textbf{Large Multimodal Models. } Recent years have witnessed the rapid growth of large multimodal models~\cite{alayrac2022flamingo,chen2023pali3,huang2023language,sun2023generative}. 
CLIP~\cite{radford2021clip} pioneered the learning of LMMs with a contrastive learning objective on massive image-text pair data. Flamingo~\cite{alayrac2022flamingo} and Kosmos~\cite{huang2023language,peng2023kosmos} exhibit promising zero-shot and few-shot multi-modal understanding performance by training on large-scale image-text interleaved data. 
With the remarkable progress in open-sourced LLMs, \cite{li2022blip,li2023blip,liu2023visual,zhu2023minigpt} show promising results by connecting vision encoders and LLMs with a small intermediate model. 
A school of successive efforts~\cite{ye2023mplugowl,zhang2023vpgtrans,su2023pandagpt,zhang2023internlmxcomposor,wang2023visionllm} further improves visual instruction tuning with better overall training pipelines~\cite{bai2023qwen,li2023otter}, grounding annotations~\cite{chen2023shikra,zhang2023gpt4roi,chen2023position,you2023ferret}, and extra tasks~\cite{bai2023qwen}. 
There are early studies on training more unified large multimodal models~\cite{sun2023generative, yu2023scaling, ge2023planting, dong2023dreamllm} that are capable of performing visual understanding and generation simultaneously.
In this paper, we further explore the distinct solution proposed in Emu~\cite{sun2023generative}: learning large multimodal models with generative objectives on both texts and images. 

\vspace{4pt} 
\noindent\textbf{In-Context Learning. } Recent advancements in large language models~\cite{brown2020language,chowdhery2022palm} underscore their capacity for in-context learning~\cite{brown2020language}. This phenomenon, particularly evident as LLMs scale up in size and data, has been exploited for complex challenges such as mathematical reasoning~\cite{wei2022chain}, signaling new emergent ability in model behavior~\cite{wei2022emergent}. 
Flamingo~\cite{alayrac2022flamingo} integrates visual inputs to LLMs, enabling the in-context learning of visual-linguistic tasks such as image captioning and OCR through language-based interfacing.  
Painter~\cite{wang2023images} and SegGPT~\cite{wang2023seggpt} conduct an early study of visual in-context learning.
Inspired by the emerging abilities of large language models, in this work we study the problem of multimodal in-context learning by scaling up generative multimodal models and demonstrating strong results in broad understanding and generation tasks.

\section{Conclusion}
We present a 37 billion-parameter generative multimodal model \Ours that shows strong performance and versatility on many multimodal tasks in the in-context settings.
\Ours serves as a base model and a general-purpose interface for a variety of multimodal tasks.
We demonstrate state-of-the-art results on a broad range of benchmarks of multimodal understanding and generation.
Specifically, our model largely surpasses prior work on the lately proposed LMM benchmarks that require more advanced capability compared to classic academic benchmarks.
\Ours also shows remarkable capability of controllable visual generation in multimodal context, \eg, subject-/text-grounded generation.
Additionally, we review the limitations and broader social impact of \Ours.
Despite discussed weaknesses, these results suggest that generative multimodal model at scale may be an important step towards the development of adaptable, general multimodal systems.

{
    \small
    \bibliographystyle{ieeenat_fullname}
    \bibliography{main}

\begin{thebibliography}{95}
\providecommand{\natexlab}[1]{#1}
\providecommand{\url}[1]{\texttt{#1}}
\expandafter\ifx\csname urlstyle\endcsname\relax
  \providecommand{\doi}[1]{doi: #1}\else
  \providecommand{\doi}{doi: \begingroup \urlstyle{rm}\Url}\fi

\bibitem[lai({\natexlab{a}})]{laionaesthetics}
Laion-aesthetics.
\newblock \url{https://laion.ai/blog/laion-aesthetics/}, {\natexlab{a}}.

\bibitem[lai({\natexlab{b}})]{laioncoco}
Laion coco: 600m synthetic captions from laion2b-en.
\newblock \url{https://laion.ai/blog/laion-coco/}, {\natexlab{b}}.

\bibitem[lai({\natexlab{c}})]{laionhighresolution}
Laion-high-resolution.
\newblock \url{https://huggingface.co/datasets/laion/laion-high-resolution}, {\natexlab{c}}.

\bibitem[sha()]{sharegpt}
Sharegpt.
\newblock \url{https://sharegpt.com/}.

\bibitem[Alayrac et~al.(2022)Alayrac, Donahue, Luc, Miech, Barr, Hasson, Lenc, Mensch, Millican, Reynolds, et~al.]{alayrac2022flamingo}
Jean-Baptiste Alayrac, Jeff Donahue, Pauline Luc, Antoine Miech, Iain Barr, Yana Hasson, Karel Lenc, Arthur Mensch, Katherine Millican, Malcolm Reynolds, et~al.
\newblock Flamingo: a visual language model for few-shot learning.
\newblock \emph{Advances in Neural Information Processing Systems}, 35:\penalty0 23716--23736, 2022.

\bibitem[Bai et~al.(2023{\natexlab{a}})Bai, Bai, Yang, Wang, Tan, Wang, Lin, Zhou, and Zhou]{bai2023qwen}
Jinze Bai, Shuai Bai, Shusheng Yang, Shijie Wang, Sinan Tan, Peng Wang, Junyang Lin, Chang Zhou, and Jingren Zhou.
\newblock Qwen-vl: A frontier large vision-language model with versatile abilities.
\newblock \emph{arXiv preprint arXiv:2308.12966}, 2023{\natexlab{a}}.

\bibitem[Bai et~al.(2023{\natexlab{b}})Bai, Yang, Bai, Wang, Zhang, Lin, Wang, Zhou, and Zhou]{bai2023touchstone}
Shuai Bai, Shusheng Yang, Jinze Bai, Peng Wang, Xingxuan Zhang, Junyang Lin, Xinggang Wang, Chang Zhou, and Jingren Zhou.
\newblock Touchstone: Evaluating vision-language models by language models.
\newblock \emph{arXiv preprint arXiv:2308.16890}, 2023{\natexlab{b}}.

\bibitem[Bain et~al.(2021)Bain, Nagrani, Varol, and Zisserman]{bain2021frozen}
Max Bain, Arsha Nagrani, G{\"u}l Varol, and Andrew Zisserman.
\newblock Frozen in time: A joint video and image encoder for end-to-end retrieval.
\newblock In \emph{Proceedings of the IEEE/CVF International Conference on Computer Vision}, pages 1728--1738, 2021.

\bibitem[Betker et~al.(2023)Betker, Goh, Jing, Brooks, Wang, Li, Ouyang, Zhuang, Lee, Guo, Manassra, Dhariwal, Chu, Jiao, and Ramesh]{betker2023dalle3}
James Betker, Gabriel Goh, Li Jing, Tim Brooks, Jianfeng Wang, Linjie Li, Long Ouyang, Juntang Zhuang, Joyce Lee, Yufei Guo, Wesam Manassra, Prafulla Dhariwal, Casey Chu, Yunxin Jiao, and Aditya Ramesh.
\newblock Improving image generation with better captions.
\newblock 2023.

\bibitem[Brooks et~al.(2023)Brooks, Holynski, and Efros]{brooks2023instructpix2pix}
Tim Brooks, Aleksander Holynski, and Alexei~A Efros.
\newblock Instructpix2pix: Learning to follow image editing instructions.
\newblock In \emph{Proceedings of the IEEE/CVF Conference on Computer Vision and Pattern Recognition}, pages 18392--18402, 2023.

\bibitem[Brown et~al.(2020)Brown, Mann, Ryder, Subbiah, Kaplan, Dhariwal, Neelakantan, Shyam, Sastry, Askell, et~al.]{brown2020language}
Tom Brown, Benjamin Mann, Nick Ryder, Melanie Subbiah, Jared~D Kaplan, Prafulla Dhariwal, Arvind Neelakantan, Pranav Shyam, Girish Sastry, Amanda Askell, et~al.
\newblock Language models are few-shot learners.
\newblock \emph{Advances in neural information processing systems}, 33:\penalty0 1877--1901, 2020.

\bibitem[Caron et~al.(2021)Caron, Touvron, Misra, J{\'e}gou, Mairal, Bojanowski, and Joulin]{caron2021emerging}
Mathilde Caron, Hugo Touvron, Ishan Misra, Herv{\'e} J{\'e}gou, Julien Mairal, Piotr Bojanowski, and Armand Joulin.
\newblock Emerging properties in self-supervised vision transformers.
\newblock In \emph{Proceedings of the IEEE/CVF international conference on computer vision}, pages 9650--9660, 2021.

\bibitem[Chang et~al.(2023)Chang, Zhang, Barber, Maschinot, Lezama, Jiang, Yang, Murphy, Freeman, Rubinstein, et~al.]{chang2023muse}
Huiwen Chang, Han Zhang, Jarred Barber, AJ Maschinot, Jose Lezama, Lu Jiang, Ming-Hsuan Yang, Kevin Murphy, William~T Freeman, Michael Rubinstein, et~al.
\newblock Muse: Text-to-image generation via masked generative transformers.
\newblock \emph{arXiv preprint arXiv:2301.00704}, 2023.

\bibitem[Chen et~al.(2023{\natexlab{a}})Chen, Qin, Luo, Mi, Li, Sun, and Liu]{chen2023position}
Chi Chen, Ruoyu Qin, Fuwen Luo, Xiaoyue Mi, Peng Li, Maosong Sun, and Yang Liu.
\newblock Position-enhanced visual instruction tuning for multimodal large language models.
\newblock \emph{arXiv preprint arXiv:2308.13437}, 2023{\natexlab{a}}.

\bibitem[Chen et~al.(2023{\natexlab{b}})Chen, Zhang, Zeng, Zhang, Zhu, and Zhao]{chen2023shikra}
Keqin Chen, Zhao Zhang, Weili Zeng, Richong Zhang, Feng Zhu, and Rui Zhao.
\newblock Shikra: Unleashing multimodal llm's referential dialogue magic.
\newblock \emph{arXiv preprint arXiv:2306.15195}, 2023{\natexlab{b}}.

\bibitem[Chen et~al.(2022)Chen, Hu, Saharia, and Cohen]{chen2022re}
Wenhu Chen, Hexiang Hu, Chitwan Saharia, and William~W Cohen.
\newblock Re-imagen: Retrieval-augmented text-to-image generator.
\newblock \emph{arXiv preprint arXiv:2209.14491}, 2022.

\bibitem[Chen et~al.(2023{\natexlab{c}})Chen, Hu, Li, Rui, Jia, Chang, and Cohen]{chen2023subject}
Wenhu Chen, Hexiang Hu, Yandong Li, Nataniel Rui, Xuhui Jia, Ming-Wei Chang, and William~W Cohen.
\newblock Subject-driven text-to-image generation via apprenticeship learning.
\newblock \emph{arXiv preprint arXiv:2304.00186}, 2023{\natexlab{c}}.

\bibitem[Chen et~al.(2015)Chen, Fang, Lin, Vedantam, Gupta, Doll{\'a}r, and Zitnick]{chen2015microsoft}
Xinlei Chen, Hao Fang, Tsung-Yi Lin, Ramakrishna Vedantam, Saurabh Gupta, Piotr Doll{\'a}r, and C~Lawrence Zitnick.
\newblock Microsoft coco captions: Data collection and evaluation server.
\newblock \emph{arXiv preprint arXiv:1504.00325}, 2015.

\bibitem[Chen et~al.(2023{\natexlab{d}})Chen, Wang, Beyer, Kolesnikov, Wu, Voigtlaender, Mustafa, Goodman, Alabdulmohsin, Padlewski, et~al.]{chen2023pali3}
Xi Chen, Xiao Wang, Lucas Beyer, Alexander Kolesnikov, Jialin Wu, Paul Voigtlaender, Basil Mustafa, Sebastian Goodman, Ibrahim Alabdulmohsin, Piotr Padlewski, et~al.
\newblock Pali-3 vision language models: Smaller, faster, stronger.
\newblock \emph{arXiv preprint arXiv:2310.09199}, 2023{\natexlab{d}}.

\bibitem[Chesser and Carbone()]{unsplash}
Luke Chesser and Timothy Carbone.
\newblock Unsplash.
\newblock \url{https://github.com/unsplash/datasets}.

\bibitem[Chowdhery et~al.(2022)Chowdhery, Narang, Devlin, Bosma, Mishra, Roberts, Barham, Chung, Sutton, Gehrmann, et~al.]{chowdhery2022palm}
Aakanksha Chowdhery, Sharan Narang, Jacob Devlin, Maarten Bosma, Gaurav Mishra, Adam Roberts, Paul Barham, Hyung~Won Chung, Charles Sutton, Sebastian Gehrmann, et~al.
\newblock Palm: Scaling language modeling with pathways.
\newblock \emph{arXiv preprint arXiv:2204.02311}, 2022.

\bibitem[Dai et~al.(2023)Dai, Li, Li, Tiong, Zhao, Wang, Li, Fung, and Hoi]{dai2023InstructBLIP}
Wenliang Dai, Junnan Li, Dongxu Li, Anthony Meng~Huat Tiong, Junqi Zhao, Weisheng Wang, Boyang~Albert Li, Pascale Fung, and Steven C.~H. Hoi.
\newblock Instructblip: Towards general-purpose vision-language models with instruction tuning.
\newblock \emph{arXiv preprint arXiv:2305.06500}, 2023.

\bibitem[Dong et~al.(2023)Dong, Han, Peng, Qi, Ge, Yang, Zhao, Sun, Zhou, Wei, Kong, Zhang, Ma, and Yi]{dong2023dreamllm}
Runpei Dong, Chunrui Han, Yuang Peng, Zekun Qi, Zheng Ge, Jinrong Yang, Liang Zhao, Jianjian Sun, Hongyu Zhou, Haoran Wei, Xiangwen Kong, Xiangyu Zhang, Kaisheng Ma, and Li Yi.
\newblock Dreamllm: Synergistic multimodal comprehension and creation.
\newblock \emph{arXiv preprint arXiv:2309.11499}, 2023.

\bibitem[Gal et~al.(2022)Gal, Alaluf, Atzmon, Patashnik, Bermano, Chechik, and Cohen-Or]{gal2022image}
Rinon Gal, Yuval Alaluf, Yuval Atzmon, Or Patashnik, Amit~H Bermano, Gal Chechik, and Daniel Cohen-Or.
\newblock An image is worth one word: Personalizing text-to-image generation using textual inversion.
\newblock \emph{arXiv preprint arXiv:2208.01618}, 2022.

\bibitem[Gan et~al.(2022)Gan, Li, Li, Wang, Liu, Gao, et~al.]{gan2022zhesurvey}
Zhe Gan, Linjie Li, Chunyuan Li, Lijuan Wang, Zicheng Liu, Jianfeng Gao, et~al.
\newblock Vision-language pre-training: Basics, recent advances, and future trends.
\newblock \emph{Foundations and Trends{\textregistered} in Computer Graphics and Vision}, 14\penalty0 (3--4):\penalty0 163--352, 2022.

\bibitem[Gao et~al.(2020)Gao, Biderman, Black, Golding, Hoppe, Foster, Phang, He, Thite, Nabeshima, Presser, and Leahy]{pile}
Leo Gao, Stella Biderman, Sid Black, Laurence Golding, Travis Hoppe, Charles Foster, Jason Phang, Horace He, Anish Thite, Noa Nabeshima, Shawn Presser, and Connor Leahy.
\newblock The {P}ile: An 800gb dataset of diverse text for language modeling.
\newblock \emph{arXiv preprint arXiv:2101.00027}, 2020.

\bibitem[Ge et~al.(2023)Ge, Ge, Zeng, Wang, and Shan]{ge2023planting}
Yuying Ge, Yixiao Ge, Ziyun Zeng, Xintao Wang, and Ying Shan.
\newblock Planting a seed of vision in large language model.
\newblock \emph{arXiv preprint arXiv:2307.08041}, 2023.

\bibitem[Goyal et~al.(2017)Goyal, Khot, Summers-Stay, Batra, and Parikh]{goyal2017making}
Yash Goyal, Tejas Khot, Douglas Summers-Stay, Dhruv Batra, and Devi Parikh.
\newblock Making the v in vqa matter: Elevating the role of image understanding in visual question answering.
\newblock In \emph{Proceedings of the IEEE conference on computer vision and pattern recognition}, pages 6904--6913, 2017.

\bibitem[Gurari et~al.(2018)Gurari, Li, Stangl, Guo, Lin, Grauman, Luo, and Bigham]{gurari2018vizwiz}
Danna Gurari, Qing Li, Abigale~J Stangl, Anhong Guo, Chi Lin, Kristen Grauman, Jiebo Luo, and Jeffrey~P Bigham.
\newblock Vizwiz grand challenge: Answering visual questions from blind people.
\newblock In \emph{Proceedings of the IEEE conference on computer vision and pattern recognition}, pages 3608--3617, 2018.

\bibitem[Ho and Salimans(2022)]{ho2022classifier}
Jonathan Ho and Tim Salimans.
\newblock Classifier-free diffusion guidance.
\newblock \emph{arXiv preprint arXiv:2207.12598}, 2022.

\bibitem[Huang et~al.(2023)Huang, Dong, Wang, Hao, Singhal, Ma, Lv, Cui, Mohammed, Liu, et~al.]{huang2023language}
Shaohan Huang, Li Dong, Wenhui Wang, Yaru Hao, Saksham Singhal, Shuming Ma, Tengchao Lv, Lei Cui, Owais~Khan Mohammed, Qiang Liu, et~al.
\newblock Language is not all you need: Aligning perception with language models.
\newblock \emph{arXiv preprint arXiv:2302.14045}, 2023.

\bibitem[Hudson and Manning(2019)]{hudson2019gqa}
Drew~A Hudson and Christopher~D Manning.
\newblock Gqa: A new dataset for real-world visual reasoning and compositional question answering.
\newblock In \emph{Proceedings of the IEEE/CVF conference on computer vision and pattern recognition}, pages 6700--6709, 2019.

\bibitem[Karras et~al.(2022)Karras, Aittala, Aila, and Laine]{karras2022elucidating}
Tero Karras, Miika Aittala, Timo Aila, and Samuli Laine.
\newblock Elucidating the design space of diffusion-based generative models.
\newblock \emph{Advances in Neural Information Processing Systems}, 35:\penalty0 26565--26577, 2022.

\bibitem[Kazemzadeh et~al.(2014)Kazemzadeh, Ordonez, Matten, and Berg]{kazemzadeh2014referitgame}
Sahar Kazemzadeh, Vicente Ordonez, Mark Matten, and Tamara Berg.
\newblock Referitgame: Referring to objects in photographs of natural scenes.
\newblock In \emph{Proceedings of the 2014 conference on empirical methods in natural language processing (EMNLP)}, pages 787--798, 2014.

\bibitem[Kiela et~al.(2020)Kiela, Firooz, Mohan, Goswami, Singh, Ringshia, and Testuggine]{kiela2020hateful}
Douwe Kiela, Hamed Firooz, Aravind Mohan, Vedanuj Goswami, Amanpreet Singh, Pratik Ringshia, and Davide Testuggine.
\newblock The hateful memes challenge: Detecting hate speech in multimodal memes.
\newblock \emph{Advances in neural information processing systems}, 33:\penalty0 2611--2624, 2020.

\bibitem[Kirillov et~al.(2023)Kirillov, Mintun, Ravi, Mao, Rolland, Gustafson, Xiao, Whitehead, Berg, Lo, et~al.]{kirillov2023segment}
Alexander Kirillov, Eric Mintun, Nikhila Ravi, Hanzi Mao, Chloe Rolland, Laura Gustafson, Tete Xiao, Spencer Whitehead, Alexander~C Berg, Wan-Yen Lo, et~al.
\newblock Segment anything.
\newblock \emph{arXiv preprint arXiv:2304.02643}, 2023.

\bibitem[Koh et~al.(2023)Koh, Fried, and Salakhutdinov]{koh2023generating}
Jing~Yu Koh, Daniel Fried, and Ruslan Salakhutdinov.
\newblock Generating images with multimodal language models.
\newblock \emph{arXiv preprint arXiv:2305.17216}, 2023.

\bibitem[Lauren{\c{c}}on et~al.(2023)Lauren{\c{c}}on, Saulnier, Tronchon, Bekman, Singh, Lozhkov, Wang, Karamcheti, Rush, Kiela, et~al.]{laurenccon2023obelics}
Hugo Lauren{\c{c}}on, Lucile Saulnier, L{\'e}o Tronchon, Stas Bekman, Amanpreet Singh, Anton Lozhkov, Thomas Wang, Siddharth Karamcheti, Alexander~M Rush, Douwe Kiela, et~al.
\newblock Obelics: An open web-scale filtered dataset of interleaved image-text documents.
\newblock In \emph{Thirty-seventh Conference on Neural Information Processing Systems Datasets and Benchmarks Track}, 2023.

\bibitem[Li et~al.(2023{\natexlab{a}})Li, Wang, Wang, Ge, Ge, and Shan]{li2023seed}
Bohao Li, Rui Wang, Guangzhi Wang, Yuying Ge, Yixiao Ge, and Ying Shan.
\newblock Seed-bench: Benchmarking multimodal llms with generative comprehension.
\newblock \emph{arXiv preprint arXiv:2307.16125}, 2023{\natexlab{a}}.

\bibitem[Li et~al.(2023{\natexlab{b}})Li, Zhang, Chen, Wang, Yang, and Liu]{li2023otter}
Bo Li, Yuanhan Zhang, Liangyu Chen, Jinghao Wang, Jingkang Yang, and Ziwei Liu.
\newblock Otter: A multi-modal model with in-context instruction tuning.
\newblock \emph{arXiv preprint arXiv:2305.03726}, 2023{\natexlab{b}}.

\bibitem[Li et~al.(2023{\natexlab{c}})Li, Gan, Yang, Yang, Li, Wang, and Gao]{li2023multimodalsurvey}
Chunyuan Li, Zhe Gan, Zhengyuan Yang, Jianwei Yang, Linjie Li, Lijuan Wang, and Jianfeng Gao.
\newblock Multimodal foundation models: From specialists to general-purpose assistants.
\newblock \emph{arXiv preprint arXiv:2309.10020}, 1\penalty0 (2):\penalty0 2, 2023{\natexlab{c}}.

\bibitem[Li et~al.(2023{\natexlab{d}})Li, Li, and Hoi]{li2023blip}
Dongxu Li, Junnan Li, and Steven~CH Hoi.
\newblock Blip-diffusion: Pre-trained subject representation for controllable text-to-image generation and editing.
\newblock \emph{arXiv preprint arXiv:2305.14720}, 2023{\natexlab{d}}.

\bibitem[Li et~al.(2022)Li, Li, Xiong, and Hoi]{li2022blip}
Junnan Li, Dongxu Li, Caiming Xiong, and Steven Hoi.
\newblock Blip: Bootstrapping language-image pre-training for unified vision-language understanding and generation.
\newblock In \emph{International Conference on Machine Learning}, pages 12888--12900. PMLR, 2022.

\bibitem[Li et~al.(2023{\natexlab{e}})Li, He, Wang, Li, Wang, Luo, Wang, Wang, and Qiao]{li2023videochat}
KunChang Li, Yinan He, Yi Wang, Yizhuo Li, Wenhai Wang, Ping Luo, Yali Wang, Limin Wang, and Yu Qiao.
\newblock Videochat: Chat-centric video understanding.
\newblock \emph{arXiv preprint arXiv:2305.06355}, 2023{\natexlab{e}}.

\bibitem[Li et~al.(2023{\natexlab{f}})Li, Yin, Li, Chen, Wang, Ren, Li, Yang, Xu, Sun, et~al.]{li2023m}
Lei Li, Yuwei Yin, Shicheng Li, Liang Chen, Peiyi Wang, Shuhuai Ren, Mukai Li, Yazheng Yang, Jingjing Xu, Xu Sun, et~al.
\newblock M3it: A large-scale dataset towards multi-modal multilingual instruction tuning.
\newblock \emph{arXiv preprint arXiv:2306.04387}, 2023{\natexlab{f}}.

\bibitem[Li et~al.(2023{\natexlab{g}})Li, Chu, Wu, Yuan, Liu, Zhang, Li, Feng, Ding, and Wang]{videogen}
Xin Li, Wenqing Chu, Ye Wu, Weihang Yuan, Fanglong Liu, Qi Zhang, Fu Li, Haocheng Feng, Errui Ding, and Jingdong Wang.
\newblock Videogen: A reference-guided latent diffusion approach for high definition text-to-video generation.
\newblock \emph{arXiv preprint arXiv:2309.00398}, 2023{\natexlab{g}}.

\bibitem[Lin et~al.(2014)Lin, Maire, Belongie, Hays, Perona, Ramanan, Doll{\'a}r, and Zitnick]{lin2014microsoft}
Tsung-Yi Lin, Michael Maire, Serge Belongie, James Hays, Pietro Perona, Deva Ramanan, Piotr Doll{\'a}r, and C~Lawrence Zitnick.
\newblock Microsoft coco: Common objects in context.
\newblock In \emph{Computer Vision--ECCV 2014: 13th European Conference, Zurich, Switzerland, September 6-12, 2014, Proceedings, Part V 13}, pages 740--755. Springer, 2014.

\bibitem[Liu et~al.(2023{\natexlab{a}})Liu, Li, Li, and Lee]{liu2023improved}
Haotian Liu, Chunyuan Li, Yuheng Li, and Yong~Jae Lee.
\newblock Improved baselines with visual instruction tuning.
\newblock \emph{arXiv preprint arXiv:2310.03744}, 2023{\natexlab{a}}.

\bibitem[Liu et~al.(2023{\natexlab{b}})Liu, Li, Wu, and Lee]{liu2023visual}
Haotian Liu, Chunyuan Li, Qingyang Wu, and Yong~Jae Lee.
\newblock Visual instruction tuning.
\newblock \emph{arXiv preprint arXiv:2304.08485}, 2023{\natexlab{b}}.

\bibitem[Loshchilov and Hutter(2017)]{loshchilov2017decoupled}
Ilya Loshchilov and Frank Hutter.
\newblock Decoupled weight decay regularization.
\newblock \emph{arXiv preprint arXiv:1711.05101}, 2017.

\bibitem[Lu et~al.(2022)Lu, Mishra, Xia, Qiu, Chang, Zhu, Tafjord, Clark, and Kalyan]{lu2022learn}
Pan Lu, Swaroop Mishra, Tanglin Xia, Liang Qiu, Kai-Wei Chang, Song-Chun Zhu, Oyvind Tafjord, Peter Clark, and Ashwin Kalyan.
\newblock Learn to explain: Multimodal reasoning via thought chains for science question answering.
\newblock \emph{Advances in Neural Information Processing Systems}, 35:\penalty0 2507--2521, 2022.

\bibitem[Mao et~al.(2016)Mao, Huang, Toshev, Camburu, Yuille, and Murphy]{mao2016generation}
Junhua Mao, Jonathan Huang, Alexander Toshev, Oana Camburu, Alan~L Yuille, and Kevin Murphy.
\newblock Generation and comprehension of unambiguous object descriptions.
\newblock In \emph{Proceedings of the IEEE conference on computer vision and pattern recognition}, pages 11--20, 2016.

\bibitem[Marino et~al.(2019)Marino, Rastegari, Farhadi, and Mottaghi]{marino2019ok}
Kenneth Marino, Mohammad Rastegari, Ali Farhadi, and Roozbeh Mottaghi.
\newblock Ok-vqa: A visual question answering benchmark requiring external knowledge.
\newblock In \emph{Proceedings of the IEEE/cvf conference on computer vision and pattern recognition}, pages 3195--3204, 2019.

\bibitem[Midjourney()]{Midjourney}
Midjourney.
\newblock Midjourney.
\newblock \url{https://www.midjourney.com}.

\bibitem[Nicoletti and Bass(2023)]{nicoletti2023humans}
Leonardo Nicoletti and Dina Bass.
\newblock Humans are biased: Generative ai is even worse.
\newblock \emph{Bloomberg Technology+ Equality. Accessed June}, 23:\penalty0 2023, 2023.

\bibitem[Pan et~al.(2023)Pan, Dong, Huang, Peng, Chen, and Wei]{pan2023kosmos}
Xichen Pan, Li Dong, Shaohan Huang, Zhiliang Peng, Wenhu Chen, and Furu Wei.
\newblock Kosmos-g: Generating images in context with multimodal large language models.
\newblock \emph{arXiv preprint arXiv:2310.02992}, 2023.

\bibitem[Peng et~al.(2023)Peng, Wang, Dong, Hao, Huang, Ma, and Wei]{peng2023kosmos}
Zhiliang Peng, Wenhui Wang, Li Dong, Yaru Hao, Shaohan Huang, Shuming Ma, and Furu Wei.
\newblock Kosmos-2: Grounding multimodal large language models to the world.
\newblock \emph{arXiv preprint arXiv:2306.14824}, 2023.

\bibitem[Podell et~al.(2023)Podell, English, Lacey, Blattmann, Dockhorn, Müller, Penna, and Rombach]{podell2023sdxl}
Dustin Podell, Zion English, Kyle Lacey, Andreas Blattmann, Tim Dockhorn, Jonas Müller, Joe Penna, and Robin Rombach.
\newblock Sdxl: Improving latent diffusion models for high-resolution image synthesis, 2023.

\bibitem[Radford et~al.(2021{\natexlab{a}})Radford, Kim, Hallacy, Ramesh, Goh, Agarwal, Sastry, Askell, Mishkin, Clark, et~al.]{radford2021clip}
Alec Radford, Jong~Wook Kim, Chris Hallacy, Aditya Ramesh, Gabriel Goh, Sandhini Agarwal, Girish Sastry, Amanda Askell, Pamela Mishkin, Jack Clark, et~al.
\newblock Learning transferable visual models from natural language supervision.
\newblock In \emph{International conference on machine learning}, pages 8748--8763. PMLR, 2021{\natexlab{a}}.

\bibitem[Radford et~al.(2021{\natexlab{b}})Radford, Kim, Hallacy, Ramesh, Goh, Agarwal, Sastry, Askell, Mishkin, Clark, et~al.]{radford2021learning}
Alec Radford, Jong~Wook Kim, Chris Hallacy, Aditya Ramesh, Gabriel Goh, Sandhini Agarwal, Girish Sastry, Amanda Askell, Pamela Mishkin, Jack Clark, et~al.
\newblock Learning transferable visual models from natural language supervision.
\newblock In \emph{International conference on machine learning}, pages 8748--8763. PMLR, 2021{\natexlab{b}}.

\bibitem[Ramesh et~al.(2022)Ramesh, Dhariwal, Nichol, Chu, and Chen]{ramesh2022hierarchical}
Aditya Ramesh, Prafulla Dhariwal, Alex Nichol, Casey Chu, and Mark Chen.
\newblock Hierarchical text-conditional image generation with clip latents.
\newblock \emph{arXiv preprint arXiv:2204.06125}, 2022.

\bibitem[Rombach et~al.(2022)Rombach, Blattmann, Lorenz, Esser, and Ommer]{rombach2022high}
Robin Rombach, Andreas Blattmann, Dominik Lorenz, Patrick Esser, and Bj{\"o}rn Ommer.
\newblock High-resolution image synthesis with latent diffusion models.
\newblock In \emph{Proceedings of the IEEE/CVF conference on computer vision and pattern recognition}, pages 10684--10695, 2022.

\bibitem[Ruiz et~al.(2023)Ruiz, Li, Jampani, Pritch, Rubinstein, and Aberman]{ruiz2023dreambooth}
Nataniel Ruiz, Yuanzhen Li, Varun Jampani, Yael Pritch, Michael Rubinstein, and Kfir Aberman.
\newblock Dreambooth: Fine tuning text-to-image diffusion models for subject-driven generation.
\newblock In \emph{Proceedings of the IEEE/CVF Conference on Computer Vision and Pattern Recognition}, pages 22500--22510, 2023.

\bibitem[Saharia et~al.(2022)Saharia, Chan, Saxena, Li, Whang, Denton, Ghasemipour, Gontijo~Lopes, Karagol~Ayan, Salimans, et~al.]{saharia2022photorealistic}
Chitwan Saharia, William Chan, Saurabh Saxena, Lala Li, Jay Whang, Emily~L Denton, Kamyar Ghasemipour, Raphael Gontijo~Lopes, Burcu Karagol~Ayan, Tim Salimans, et~al.
\newblock Photorealistic text-to-image diffusion models with deep language understanding.
\newblock \emph{Advances in Neural Information Processing Systems}, 35:\penalty0 36479--36494, 2022.

\bibitem[Schuhmann et~al.(2022)Schuhmann, Beaumont, Vencu, Gordon, Wightman, Cherti, Coombes, Katta, Mullis, Wortsman, et~al.]{schuhmann2022laion}
Christoph Schuhmann, Romain Beaumont, Richard Vencu, Cade Gordon, Ross Wightman, Mehdi Cherti, Theo Coombes, Aarush Katta, Clayton Mullis, Mitchell Wortsman, et~al.
\newblock Laion-5b: An open large-scale dataset for training next generation image-text models.
\newblock \emph{arXiv preprint arXiv:2210.08402}, 2022.

\bibitem[Sidorov et~al.(2020)Sidorov, Hu, Rohrbach, and Singh]{sidorov2020textcaps}
Oleksii Sidorov, Ronghang Hu, Marcus Rohrbach, and Amanpreet Singh.
\newblock Textcaps: a dataset for image captioning with reading comprehension.
\newblock In \emph{Computer Vision--ECCV 2020: 16th European Conference, Glasgow, UK, August 23--28, 2020, Proceedings, Part II 16}, pages 742--758. Springer, 2020.

\bibitem[Singer et~al.(2022)Singer, Polyak, Hayes, Yin, An, Zhang, Hu, Yang, Ashual, Gafni, et~al.]{make-a-video}
Uriel Singer, Adam Polyak, Thomas Hayes, Xi Yin, Jie An, Songyang Zhang, Qiyuan Hu, Harry Yang, Oron Ashual, Oran Gafni, et~al.
\newblock Make-a-video: Text-to-video generation without text-video data.
\newblock \emph{arXiv preprint arXiv:2209.14792}, 2022.

\bibitem[Singh et~al.(2019)Singh, Natarajan, Shah, Jiang, Chen, Batra, Parikh, and Rohrbach]{singh2019towards}
Amanpreet Singh, Vivek Natarajan, Meet Shah, Yu Jiang, Xinlei Chen, Dhruv Batra, Devi Parikh, and Marcus Rohrbach.
\newblock Towards vqa models that can read.
\newblock In \emph{Proceedings of the IEEE/CVF conference on computer vision and pattern recognition}, pages 8317--8326, 2019.

\bibitem[Su et~al.(2023)Su, Lan, Li, Xu, Wang, and Cai]{su2023pandagpt}
Yixuan Su, Tian Lan, Huayang Li, Jialu Xu, Yan Wang, and Deng Cai.
\newblock Pandagpt: One model to instruction-follow them all.
\newblock \emph{arXiv preprint arXiv:2305.16355}, 2023.

\bibitem[Sun et~al.(2023{\natexlab{a}})Sun, Fang, Wu, Wang, and Cao]{sun2023evaclip}
Quan Sun, Yuxin Fang, Ledell Wu, Xinlong Wang, and Yue Cao.
\newblock Eva-clip: Improved training techniques for clip at scale.
\newblock \emph{arXiv preprint arXiv:2303.15389}, 2023{\natexlab{a}}.

\bibitem[Sun et~al.(2023{\natexlab{b}})Sun, Yu, Cui, Zhang, Zhang, Wang, Gao, Liu, Huang, and Wang]{sun2023generative}
Quan Sun, Qiying Yu, Yufeng Cui, Fan Zhang, Xiaosong Zhang, Yueze Wang, Hongcheng Gao, Jingjing Liu, Tiejun Huang, and Xinlong Wang.
\newblock Generative pretraining in multimodality.
\newblock \emph{arXiv preprint arXiv:2307.05222}, 2023{\natexlab{b}}.

\bibitem[Taori et~al.(2023)Taori, Gulrajani, Zhang, Dubois, Li, Guestrin, Liang, and Hashimoto]{taori2023stanford}
Rohan Taori, Ishaan Gulrajani, Tianyi Zhang, Yann Dubois, Xuechen Li, Carlos Guestrin, Percy Liang, and Tatsunori~B Hashimoto.
\newblock Stanford alpaca: An instruction-following llama model, 2023.

\bibitem[Touvron et~al.(2023)Touvron, Lavril, Izacard, Martinet, Lachaux, Lacroix, Rozière, Goyal, Hambro, Azhar, Rodriguez, Joulin, Grave, and Lample]{touvron2023llama}
Hugo Touvron, Thibaut Lavril, Gautier Izacard, Xavier Martinet, Marie-Anne Lachaux, Timothée Lacroix, Baptiste Rozière, Naman Goyal, Eric Hambro, Faisal Azhar, Aurelien Rodriguez, Armand Joulin, Edouard Grave, and Guillaume Lample.
\newblock Llama: Open and efficient foundation language models.
\newblock \emph{arXiv preprint arXiv:2302.13971}, 2023.

\bibitem[Wang et~al.(2023{\natexlab{a}})Wang, Yuan, Chen, Zhang, Wang, and Zhang]{modelscope}
Jiuniu Wang, Hangjie Yuan, Dayou Chen, Yingya Zhang, Xiang Wang, and Shiwei Zhang.
\newblock Modelscope text-to-video technical report.
\newblock \emph{arXiv preprint arXiv:2308.06571}, 2023{\natexlab{a}}.

\bibitem[Wang et~al.(2022)Wang, Yang, Men, Lin, Bai, Li, Ma, Zhou, Zhou, and Yang]{wang2022ofa}
Peng Wang, An Yang, Rui Men, Junyang Lin, Shuai Bai, Zhikang Li, Jianxin Ma, Chang Zhou, Jingren Zhou, and Hongxia Yang.
\newblock Ofa: Unifying architectures, tasks, and modalities through a simple sequence-to-sequence learning framework.
\newblock In \emph{International Conference on Machine Learning}, pages 23318--23340. PMLR, 2022.

\bibitem[Wang et~al.(2023{\natexlab{b}})Wang, Chen, Chen, Wu, Zhu, Zeng, Luo, Lu, Zhou, Qiao, et~al.]{wang2023visionllm}
Wenhai Wang, Zhe Chen, Xiaokang Chen, Jiannan Wu, Xizhou Zhu, Gang Zeng, Ping Luo, Tong Lu, Jie Zhou, Yu Qiao, et~al.
\newblock Visionllm: Large language model is also an open-ended decoder for vision-centric tasks.
\newblock \emph{arXiv preprint arXiv:2305.11175}, 2023{\natexlab{b}}.

\bibitem[Wang et~al.(2023{\natexlab{c}})Wang, Lv, Yu, Hong, Qi, Wang, Ji, Yang, Zhao, Song, et~al.]{wang2023cogvlm}
Weihan Wang, Qingsong Lv, Wenmeng Yu, Wenyi Hong, Ji Qi, Yan Wang, Junhui Ji, Zhuoyi Yang, Lei Zhao, Xixuan Song, et~al.
\newblock Cogvlm: Visual expert for pretrained language models.
\newblock \emph{arXiv preprint arXiv:2311.03079}, 2023{\natexlab{c}}.

\bibitem[Wang et~al.(2023{\natexlab{d}})Wang, Wang, Cao, Shen, and Huang]{wang2023images}
Xinlong Wang, Wen Wang, Yue Cao, Chunhua Shen, and Tiejun Huang.
\newblock Images speak in images: A generalist painter for in-context visual learning.
\newblock In \emph{Proceedings of the IEEE/CVF Conference on Computer Vision and Pattern Recognition}, pages 6830--6839, 2023{\natexlab{d}}.

\bibitem[Wang et~al.(2023{\natexlab{e}})Wang, Zhang, Cao, Wang, Shen, and Huang]{wang2023seggpt}
Xinlong Wang, Xiaosong Zhang, Yue Cao, Wen Wang, Chunhua Shen, and Tiejun Huang.
\newblock Seggpt: Segmenting everything in context.
\newblock \emph{arXiv preprint arXiv:2304.03284}, 2023{\natexlab{e}}.

\bibitem[Wei et~al.(2022{\natexlab{a}})Wei, Tay, Bommasani, Raffel, Zoph, Borgeaud, Yogatama, Bosma, Zhou, Metzler, et~al.]{wei2022emergent}
Jason Wei, Yi Tay, Rishi Bommasani, Colin Raffel, Barret Zoph, Sebastian Borgeaud, Dani Yogatama, Maarten Bosma, Denny Zhou, Donald Metzler, et~al.
\newblock Emergent abilities of large language models.
\newblock \emph{arXiv preprint arXiv:2206.07682}, 2022{\natexlab{a}}.

\bibitem[Wei et~al.(2022{\natexlab{b}})Wei, Wang, Schuurmans, Bosma, Xia, Chi, Le, Zhou, et~al.]{wei2022chain}
Jason Wei, Xuezhi Wang, Dale Schuurmans, Maarten Bosma, Fei Xia, Ed Chi, Quoc~V Le, Denny Zhou, et~al.
\newblock Chain-of-thought prompting elicits reasoning in large language models.
\newblock \emph{Advances in Neural Information Processing Systems}, 35:\penalty0 24824--24837, 2022{\natexlab{b}}.

\bibitem[Xu et~al.(2017)Xu, Zhao, Xiao, Wu, Zhang, He, and Zhuang]{xu2017video}
Dejing Xu, Zhou Zhao, Jun Xiao, Fei Wu, Hanwang Zhang, Xiangnan He, and Yueting Zhuang.
\newblock Video question answering via gradually refined attention over appearance and motion.
\newblock In \emph{Proceedings of the 25th ACM international conference on Multimedia}, pages 1645--1653, 2017.

\bibitem[Yang et~al.(2022)Yang, Xie, and Zisserman]{yang2022s}
Charig Yang, Weidi Xie, and Andrew Zisserman.
\newblock It's about time: Analog clock reading in the wild.
\newblock In \emph{Proceedings of the IEEE/CVF Conference on Computer Vision and Pattern Recognition}, pages 2508--2517, 2022.

\bibitem[Ye et~al.(2023)Ye, Xu, Xu, Ye, Yan, Zhou, Wang, Hu, Shi, Shi, et~al.]{ye2023mplugowl}
Qinghao Ye, Haiyang Xu, Guohai Xu, Jiabo Ye, Ming Yan, Yiyang Zhou, Junyang Wang, Anwen Hu, Pengcheng Shi, Yaya Shi, et~al.
\newblock mplug-owl: Modularization empowers large language models with multimodality.
\newblock \emph{arXiv preprint arXiv:2304.14178}, 2023.

\bibitem[You et~al.(2023)You, Zhang, Gan, Du, Zhang, Wang, Cao, Chang, and Yang]{you2023ferret}
Haoxuan You, Haotian Zhang, Zhe Gan, Xianzhi Du, Bowen Zhang, Zirui Wang, Liangliang Cao, Shih-Fu Chang, and Yinfei Yang.
\newblock Ferret: Refer and ground anything anywhere at any granularity.
\newblock \emph{arXiv preprint arXiv:2310.07704}, 2023.

\bibitem[Yu et~al.(2023{\natexlab{a}})Yu, Shi, Pasunuru, Muller, Golovneva, Wang, Babu, Tang, Karrer, Sheynin, et~al.]{yu2023scaling}
Lili Yu, Bowen Shi, Ramakanth Pasunuru, Benjamin Muller, Olga Golovneva, Tianlu Wang, Arun Babu, Binh Tang, Brian Karrer, Shelly Sheynin, et~al.
\newblock Scaling autoregressive multi-modal models: Pretraining and instruction tuning.
\newblock \emph{arXiv preprint arXiv:2309.02591}, 2023{\natexlab{a}}.

\bibitem[Yu et~al.(2023{\natexlab{b}})Yu, Sun, Zhang, Cui, Zhang, Wang, and Liu]{yu2023capsfusion}
Qiying Yu, Quan Sun, Xiaosong Zhang, Yufeng Cui, Fan Zhang, Xinlong Wang, and Jingjing Liu.
\newblock Capsfusion: Rethinking image-text data at scale.
\newblock \emph{arXiv preprint arXiv:2310.20550}, 2023{\natexlab{b}}.

\bibitem[Yu et~al.(2023{\natexlab{c}})Yu, Yang, Li, Wang, Lin, Liu, Wang, and Wang]{yu2023mm}
Weihao Yu, Zhengyuan Yang, Linjie Li, Jianfeng Wang, Kevin Lin, Zicheng Liu, Xinchao Wang, and Lijuan Wang.
\newblock Mm-vet: Evaluating large multimodal models for integrated capabilities.
\newblock \emph{arXiv preprint arXiv:2308.02490}, 2023{\natexlab{c}}.

\bibitem[Yue et~al.(2023)Yue, Ni, Zhang, Zheng, Liu, Zhang, Stevens, Jiang, Ren, Sun, et~al.]{yue2023mmmu}
Xiang Yue, Yuansheng Ni, Kai Zhang, Tianyu Zheng, Ruoqi Liu, Ge Zhang, Samuel Stevens, Dongfu Jiang, Weiming Ren, Yuxuan Sun, et~al.
\newblock Mmmu: A massive multi-discipline multimodal understanding and reasoning benchmark for expert agi.
\newblock \emph{arXiv preprint arXiv:2311.16502}, 2023.

\bibitem[Zhang et~al.(2023{\natexlab{a}})Zhang, Fei, Yao, Ji, Li, Liu, and Chua]{zhang2023vpgtrans}
Ao Zhang, Hao Fei, Yuan Yao, Wei Ji, Li Li, Zhiyuan Liu, and Tat-Seng Chua.
\newblock Transfer visual prompt generator across llms.
\newblock \emph{arXiv preprint arXiv:2305.01278}, 2023{\natexlab{a}}.

\bibitem[Zhang et~al.(2023{\natexlab{b}})Zhang, Wang, Cao, Xu, Ouyang, Zhao, Ding, Zhang, Duan, Yan, et~al.]{zhang2023internlmxcomposor}
Pan Zhang, Xiaoyi Dong~Bin Wang, Yuhang Cao, Chao Xu, Linke Ouyang, Zhiyuan Zhao, Shuangrui Ding, Songyang Zhang, Haodong Duan, Hang Yan, et~al.
\newblock Internlm-xcomposer: A vision-language large model for advanced text-image comprehension and composition.
\newblock \emph{arXiv preprint arXiv:2309.15112}, 2023{\natexlab{b}}.

\bibitem[Zhang et~al.(2023{\natexlab{c}})Zhang, Sun, Chen, Xiao, Shao, Zhang, Chen, and Luo]{zhang2023gpt4roi}
Shilong Zhang, Peize Sun, Shoufa Chen, Min Xiao, Wenqi Shao, Wenwei Zhang, Kai Chen, and Ping Luo.
\newblock Gpt4roi: Instruction tuning large language model on region-of-interest.
\newblock \emph{arXiv preprint arXiv:2307.03601}, 2023{\natexlab{c}}.

\bibitem[Zhang et~al.(2023{\natexlab{d}})Zhang, Zhang, Gu, Zhou, Lipka, Yang, and Sun]{zhang2023llavar}
Yanzhe Zhang, Ruiyi Zhang, Jiuxiang Gu, Yufan Zhou, Nedim Lipka, Diyi Yang, and Tong Sun.
\newblock Llavar: Enhanced visual instruction tuning for text-rich image understanding.
\newblock \emph{arXiv preprint arXiv:2306.17107}, 2023{\natexlab{d}}.

\bibitem[Zhu et~al.(2023{\natexlab{a}})Zhu, Chen, Shen, Li, and Elhoseiny]{zhu2023minigpt}
Deyao Zhu, Jun Chen, Xiaoqian Shen, Xiang Li, and Mohamed Elhoseiny.
\newblock Minigpt-4: Enhancing vision-language understanding with advanced large language models.
\newblock \emph{arXiv preprint arXiv:2304.10592}, 2023{\natexlab{a}}.

\bibitem[Zhu et~al.(2023{\natexlab{b}})Zhu, Hessel, Awadalla, Gadre, Dodge, Fang, Yu, Schmidt, Wang, and Choi]{mmc4}
Wanrong Zhu, Jack Hessel, Anas Awadalla, Samir~Yitzhak Gadre, Jesse Dodge, Alex Fang, Youngjae Yu, Ludwig Schmidt, William~Yang Wang, and Yejin Choi.
\newblock Multimodal c4: An open, billion-scale corpus of images interleaved with text.
\newblock \emph{arXiv preprint arXiv:2304.06939}, 2023{\natexlab{b}}.

\end{thebibliography}
}

\clearpage
\setcounter{page}{1}
\maketitlesupplementary

\appendix

\section{Broader Impact and Limitations}

Large multimodal models offer a wide range of benefits to society, from enhancing visual navigation and medical diagnostics to increasing accessibility for individuals with visual impairment. 
The in-context learning capabilities of \Ours allow it to quickly adapt to new tasks or environments, even with limited data, ushering in numerous potential applications. 
The generative capabilities of \Ours can be highly valuable to the creative industries. 

However, there are potential downsides in more powerful multimodal models to be considered.   
The hallucination issue of multimodal models may cause incorrect and unreasonable predictions in certain cases. \Ours may also generate harmful or biased content like other generative models~\cite{nicoletti2023humans} since the training data may be biased or contain unsuitable content. 
We are actively working to enhance the robustness of multimodal models, reduce model hallucinations,  improve the fairness of training data, and reduce toxic data. We also call on the wider community to pay attention to the potential social impact of multimodal models as they are growing larger and stronger.

One of the limitations of \Ours is that its in-context learning capability could fail in some complex scenes or tasks, \eg, counting in a crowd. 
Additionally, there is still a gap between \Ours's question-answering capability and that of closed multimodal systems. For example, GPT-4V achieves 67.7 MM-Vet score vs.\ \Ours's 48.5, although already being state-of-the-art among public models.
We believe there is much room to improve as the quality and quantity of training data improve and as model scale continues to grow.

\section{More Pretraining Details}

\subsection{Dataset Details}

In pretraining, we exclusively leverage image-text pairs and video-text pairs for stage 1 training. We additionally leverage interleaved and language-only data altogether for stage 2. The integration of visual embeddings with text tokens generates unified multimodal sequences. These sequences are then structured by appending the tokens \texttt{<s>} and \texttt{</s>} to denote the beginning and end of each sequence.

\paragraph{Image/Video-text Pairs.}  In the pretraining stage, we utilize image-text pairs from LAION-2B~\citep{schuhmann2022laion} and CapsFusion-120M~\cite{yu2023capsfusion}, along with video-text pairs from WebVid-10M~\citep{bain2021frozen}. 
During pretraining stage 2, each image or video is randomly placed before or after its corresponding text with a probability of 0.5, respectively. For each video, we randomly sample 8 frames. To structure the visual embeddings, we append two special tokens, \texttt{[IMG]} and \texttt{[/IMG]}, to signify the start and end of the visual embeddings. In the case of videos, where there are $T$ frames, each frame is encoded into a set of visual embeddings, and a special token, \texttt{[VIDEO]}, is prepended to the start of the frame embedding sequence. This design helps distinguish between multiple images and video frames within the multimodal sequences.

\paragraph{Interleaved Image/Video-text Data.} We harness the Multimodal-C4 (MMC4) dataset~\citep{mmc4} and the YT-Storyboard-1B dataset~\citep{sun2023generative} as expansive sources of image and video-text interleaved data. This approach aims to unlock the in-context learning capability of multimodal models. For each MMC4 document, we randomly sample N = 8 images, accompanied by their corresponding sentences, to construct a subsequence of L = 1024. During pretraining stage 2, each image or frame is randomly positioned before or after its corresponding text with a probability of 0.5. The special tokens used in this interleaved data are consistent with those employed in the image-text pair data.

\paragraph{Grounded Image-text Pairs.} We curated a dataset of grounded image-text pairs named CapsFusion-grounded-100M, employing data from CapsFusion~\cite{yu2023capsfusion} processed through the dataset construction pipeline proposed by Kosmos-2~\citep{peng2023kosmos}. Additionally, we utilized the 20M GRIT dataset introduced by Kosmos-2~\citep{peng2023kosmos}. To enhance the diversity and context of the dataset, we randomly positioned each phrase before or after its corresponding coordinates with a probability of 0.7. The bounding box can be represented using its top-left point $(x1, y1)$ and bottom-right point $(x2, y2)$. We transform continuous coordinates into 224 discrete tokens~\citep{peng2023kosmos}, the coordinates of a sample box can be formulated as \texttt{<$loc_{000}$>}\texttt{<$loc_{000}$>}\texttt{<$loc_{224}$>}\texttt{<$loc_{224}$>}. We added these tokens to the word vocabulary to facilitate unified modeling with text. 
To distinguish grounding text from regular text strings, we introduced two special tokens, \texttt{<coor>} and \texttt{</coor>}, marking the beginning and end of the bounding box coordinates. Moreover, to establish the correct association between bounding boxes and their corresponding descriptive phrases, an additional set of special tokens, \texttt{<p>} and \texttt{</p>}, was appended. To guide the model in grounding text output to the provided image, we utilized the special token \texttt{<grounding>}. 
This comprehensive set of tokens and instructions enriches the training data for effective multimodal modeling and understanding.

\paragraph{Language-only Data.} To maintain text reasoning capabilities, we engage in joint training with the language modeling dataset Pile~\citep{pile}. The entire text corpus from Pile is preprocessed offline, and each training sample is tokenized into 2048 tokens using the LLaMA tokenizer. We randomly sample a total of 3.6 billion tokens for pretraining purposes.

\subsection{Training Hyperparameters}
We report the detailed training hyperparameter settings of \Ours during the pretraining in Table~\ref{tab:emu pretraining cfg}.

\begin{table}[h]
    \centering
    \setlength{\tabcolsep}{0.15cm}
    \renewcommand{\arraystretch}{1.2}
    \resizebox{1.0\linewidth}{!}{
    \begin{tabular}{l cc}
         \toprule
         Configuration   & \Ours Stage 1 & \Ours Stage 2 \\
         \midrule
         Visual Encoder init.   & EVA-02-CLIP-E-plus & \Ours stage 1 \\
         Multimodel Modeling init.                & LLaMA-33B & \Ours stage 1 \\
         Linear projection layer init.         & random & \Ours stage 1 \\
         Input image resolution         & $224^2$ \/ $448^2$ & $448^2$ \\
         Optimizer                & \multicolumn{2}{c}{AdamW} \\
         Optimizer hyper-parameters & \multicolumn{2}{c}{$\beta_{1}=0.9, \beta_{2}=0.95, eps=10^{-6}$} \\
         Peak learning rate       & \(1 \times 10^{-4}\), \(3 \times 10^{-5}\), \(5 \times 10^{-5}\)  & \(1 \times 10^{-5}\) \\
         Learning rate schedule   & \multicolumn{2}{c}{cosine decay} \\
         Gradient clip            & \multicolumn{2}{c}{5.0} \\
         Training steps           & 35.2k \/ 4.0k & 20.35k \\
         Warmup ratio & 0.02 & 0.1 \\
         Global batch size*        & 6144, 768 & 12800, 6400, 3200, 800\\
         Numerical precision      & \multicolumn{2}{c}{$\mathtt{bfloat16}$} \\
         \bottomrule
    \end{tabular}
  }
  \caption{Summary of pretraining hyperparameters of \Ours in pretraining stages. Peaking leaning rates are \(1 \times 10^{-4}\) for the linear projection layer, \(3 \times 10^{-5}\) for Multimodel Modeling, and \(5 \times 10^{-5}\) for Visual Encoder. *Global batch size: 1) 6144 for image-text pairs and 768 for video-text pairs in stage 1. 2) 12800 for image-text pairs, 6400 for video-text pairs, 3200 for image-text/video-text interleaved data, and 800 for language-only data in stage 2.}
  \label{tab:emu pretraining cfg}
\end{table}

\subsection{Visual Decoding}
\label{app:visual_decoding}

\subsubsection{Dataset Details}

We utilize images in LAION-COCO~\citep{laioncoco} and LAION-Aesthetics~\citep{laionaesthetics} to train the Visual Decoder. Images whose resolution is smaller than $512 \times 512$ are filtered to prevent generating low-quality results. We employ ratio-preserving random scaling followed by random cropping of a square portion from the scaled image to keep all training images unstretched. The original image size and crop coordinates are used as additional conditions following SDXL~\citep{podell2023sdxl}.

\subsubsection{Training Hyperparameters}

The detailed hyperparameters of visual decoding training are summarized in Table \ref{tab:visual_decoding_train_setting}.

\begin{table}[h]
    \centering
    \setlength{\tabcolsep}{0.15cm}
    \renewcommand{\arraystretch}{1.2}
    \resizebox{0.85\linewidth}{!}{
    \begin{tabular}{l c}
         \toprule
         Configuration                      & Visual Decoding  \\
         \midrule
         Visual Encoder init.               & \Ours stage 1 \\
         Visual Decoder init.               & SDXL-base \\
         Encoder input image resolution     & $448 \times 448$ \\
         Decoder output image resolution    & $1024 \times 1024$ \\
         Optimizer                          & AdamW \\
         Optimizer hyper-parameters         & $\beta_{1}=0.9, \beta_{2}=0.999, eps=10^{-8}$ \\
         Peak learning rate                 & $1 \times 10^{-4}$ \\
         Learning rate schedule             & $log$ warm-up, linear decay \\
         Gradient clip                      & 1.0 \\
         Total training steps               & 8,000 \\
         Warmup steps                       & 2,500 \\
         batch size                         & 2,048 \\
         Numerical precision                & $\mathtt{bfloat16}$ \\
         Classifier-free guidance           & 10\% \\
         Noise offset                       & 0.1 \\
         \bottomrule
    \end{tabular}
  }
  \caption{Summary of training hyperparameters of \Ours Visual Decoder. The Visual Encoder is frozen during training.}
  \label{tab:visual_decoding_train_setting}
\end{table}

\section{Instruction-Following Chat}

\subsection{Dataset Details}

We used two types of training data, academic-task-oriented data and multi-modal chat data, in instruction fine-tuning of \OursChat
The academic-task-oriented datasets we utilized comprise image captioning datasets such as COCO Caption~\citep{chen2015microsoft}, and TextCaps~\citep{sidorov2020textcaps}, as well as visual question-answering datasets like VQAv2~\citep{goyal2017making}, OKVQA~\citep{marino2019ok}, GQA~\citep{hudson2019gqa}, TextVQA~\citep{singh2019towards}, and multi-modal classification data constructed in M3IT~\citep{li2023m}. RefCOCO~\citep{kazemzadeh2014referitgame}, RefCOCO+~\citep{mao2016generation} and RefCOCOg~\citep{mao2016generation} datasets are also used. The public multi-modal chat data we use includes GPT-assisted visual instruction data LLaVa~\citep{liu2023visual} and LLaVaR~\citep{zhang2023llavar}, language instruction data from ShareGPT~\citep{sharegpt} and Alpaca~\citep{taori2023stanford}, and video instruction data from VideoChat~\citep{li2023videochat}. Beyond these, we constructed instruction fine-tuning data from an analog clock reading dataset~\citep{yang2022s}. For academic-task-oriented datasets, we use the system message ``\prompt{You are a helpful assistant, dedicated to provide concise and efficient answers.}'', and for the multi-modal chat data, the system message is ``\prompt{You are a helpful assistant, dedicated to delivering comprehensive and meticulous responses.}''.

\subsection{Training Hyperparameters}

The detailed training hyper-parameters of \OursChat are summarized in Table~\ref{tabsupp:chatconfig}. 

\begin{table}[h]
    \centering
    \setlength{\tabcolsep}{0.15cm}
    \renewcommand{\arraystretch}{1.2}
    \resizebox{0.85\linewidth}{!}{
    \begin{tabular}{l c}
         \toprule
         Configuration               
         & \OursChat                                    \\
         \midrule
         init.                       
         & \Ours                                        \\
         Input image resolution      
         & $448\times448$ \\
         Optimizer                   
         & AdamW                                        \\
         Optimizer hyper-parameters 
         & $\beta_{1}=0.9, \beta_{2}=0.98, eps=10^{-6}$ \\
         Peak learning rate       
         & $1 \times 10^{-5}$                                    \\
         Learning rate schedule   
         & cosine decay                                 \\
         Gradient clip            
         & 5.0                                          \\
         Training steps           
         & 8,000                                           \\
         Warmup steps 
         & 100                                         \\
         Global Batch size
         & 768                                          \\
         Numerical precision      
         & $\mathtt{bfloat16}$                          \\
         \bottomrule 
    \end{tabular}
  }
  \caption{Summary of training hyperparameters of \OursChat.}
  \label{tabsupp:chatconfig}
\end{table}

\section{Controllable Visual Generation}

\subsection{Dataset Details}

We use the grounded image-text pairs dataset, \ie, CapsFusion-grounded-100M and GRIT~\citep{peng2023kosmos} for grounded text-to-image generation. We use SAM~\citep{kirillov2023segment} to obtain segmentation results for the corresponding grounding boxes. 
We leverage InstructPix2Pix constructed by \citep{brooks2023instructpix2pix} for image editing tasks. The sample will be formulated as ``{\prompt{<s>[IMG]embedding of origin image[/IMG]instruct editing prompt[IMG]embedding of edited image[/IMG]</s>}}''. 
For the text-to-image task, we use a filtered subset the CapsFusion~\citep{yu2023capsfusion}, LAION-Aesthetics~\citep{laionaesthetics}, SA-1B~\citep{kirillov2023segment}, and LAION-High-Resolution~\citep{laionhighresolution}.

For high-quality fine-tuning, our datasets were meticulously sourced from premium sources, \eg, Unsplash~\citep{unsplash}, and outputs from advanced text-to-image systems, \eg, Midjourney-V5~\citep{Midjourney} and DALL-E-3~\citep{betker2023dalle3}. This comprehensive approach ensured a diverse and rich dataset, comprising approximately 500,000 instances of high-quality image-text pairs, instrumental in refining and enhancing the aesthetic quality of our \OursG model's generated images.

\subsection{Training Hyperparameters}

We report the detailed training hyperparameter settings of \OursG during the instruction-tuning in Table~\ref{tab:emug training cfg}.

\begin{table}[h]
    \centering
    \setlength{\tabcolsep}{0.15cm}
    \renewcommand{\arraystretch}{1.2}
    \resizebox{1.0\linewidth}{!}{
    \begin{tabular}{l cc}
         \toprule
         Configuration   &  \OursG stage1 & \OursG QFT \\
         \midrule
         init.   & \Ours & \OursG stage1 \\
         Input image resolution     & \multicolumn{2}{c}{$448\times448$} \\
         Optimizer                & \multicolumn{2}{c}{AdamW} \\
         Optimizer hyper-parameters & \multicolumn{2}{c}{$\beta_{1}=0.9, \beta_{2}=0.95, eps=10^{-6}$} \\
         Peak learning rate       & $5 \times 10^{-5}$  & $1 \times 10^{-5}$ \\
         Learning rate schedule   & \multicolumn{2}{c}{cosine decay} \\
         Gradient clip            & \multicolumn{2}{c}{1.0} \\
         Training steps           & 3k & 0.9k \\
         Warmup ratio & \multicolumn{2}{c}{0.0} \\
         Global Batch size* & 4096, 3584, 2048 & 2048, 1024, 2048 \\
         
         Numerical precision      & \multicolumn{2}{c}{$\mathtt{bfloat16}$} \\
         \bottomrule 
    \end{tabular}
  }
  \caption{Summary of training hyperparameters of \OursG. *Dataset types are text-to-image pairs, grounded text-to-image and image editing pairs.}
  \label{tab:emug training cfg}
\end{table}

\vspace{-15pt}
\section{Evaluation Details}

\paragraph{Pretrained Base Model.}
For few-shot evaluation of \Ours, we adopt the Retrieval In-Context Example Selection (RICES) approach for choosing few-shot examples, following Flamingo~\cite{alayrac2022flamingo} and Emu~\cite{sun2023generative}. 
The chosen few-shot examples will be separated by ``\prompt{. }'' and then placed ahead of the test sample. 
We use the prompt "\prompt{[image] based on the picture, [question] short answer:}". For zero-shot evaluation, as no example is given, we find the above simple prompt cannot effectively control the model behavior and the model tends to output a sentence rather than a word or phrase. Thus, we modify the prompt to "\prompt{[image] based on the picture, answer in one word or phrase. [question] short answer:}". This adjustment aligns the model's output more closely with the distribution of the tested datasets, where responses typically consist of a succinct word or phrase.
The splits and metrics for each benchmark are detailed in Table~\ref{tabsupp:benchmark}.

\paragraph{Instruction-Following Chat.}
The evaluation of \OursChat follows the assessment method of Emu-I~\cite{sun2023generative}, utilizing generation hyper-parameters with a beam size of 5. For video input, 16 frames are uniformly sampled as visual conditions. In the question-answering benchmark that requires short answers, we employ the system message ``\prompt{You are a helpful assistant, dedicated to provide concise and efficient answers.}'' along with the output format control information used in \cite{liu2023improved}. In the benchmark for scoring with GPT-4, we use the system message ``\prompt{You are a helpful assistant, dedicated to delivering comprehensive and meticulous responses.}''. We provide an overview of the evaluation benchmarks in Table~\ref{tabsupp:benchmark}.

\begin{table}[htbp]
  \centering
  \setlength{\tabcolsep}{0.10cm}
  \resizebox{1.0\linewidth}{!}{
  \begin{tabular}{llll}
    \toprule
    Benchmark   & Task   & Split   & Metric             \\
    \midrule
      VQAv2     & Scene understanding VQA
    & Test-dev  & VQA score($\uparrow$)          \\
      VizWiz    & Scene understanding VQA        
    & Test-dev  & VQA score($\uparrow$)          \\
      GQA       & Understanding \& reasoning VQA
    & Test-dev  & EM($\uparrow$)                 \\
      OKVQA     & External knowledge VQA         
    & Val       & VQA score($\uparrow$)          \\
      TextVQA   & Text-oriented VQA
    & Val       & VQA score($\uparrow$)          \\
      Hateful Memes & Meme classification
    & Seen Test & ROC AUC($\uparrow$)            \\
    \midrule
      RefCOCO   & Refer expression comprehension 
    & -         & Accuracy($\uparrow$)           \\
      RefCOCO+  & Refer expression comprehension 
    & -         & Accuracy($\uparrow$)           \\
      RefCOCOg  & Refer expression comprehension 
    & -         & Accuracy($\uparrow$)           \\
    \midrule
      MSVD-QA   & Event understanding VQA        
    & Test      & EM($\uparrow$)                 \\
      MSRVTT-QA & Event understanding VQA        
    & Test      & EM($\uparrow$)                 \\
    \midrule
      MMMU      & Massive multi-discipline QA
    & Test      & Accuracy($\uparrow$)           \\
      SEED-Bench& Image/Video multi-choice QA
    & -         & Accuracy($\uparrow$)           \\
      MM-Vet    & Open-ended generation
    & -         & GPT-4 score($\uparrow$)        \\
      TouchStone& Open-ended generation
    & -         & GPT-4 score($\uparrow$)        \\
    \bottomrule
  \end{tabular}
  }
  \caption{Summary of the evaluation benchmarks.}
  \label{tabsupp:benchmark}
\end{table}

\vspace{-20pt}
\paragraph{Controllable Visual Generation.}
For all evaluation of visual generation tasks, we use EulerDiscreteScheduler~\citep{karras2022elucidating} with 50 diffusion steps. The classifier-free guidance scale is set to 3.0.
To evaluate on DreamBench~\citep{ruiz2023dreambooth}, we select exactly the same image for each object as chosen in Kosmos-G~\citep{pan2023kosmos}. Similarly to Kosmos-G, we also slightly modified the original prompt with the prefix "a" , for example, "\prompt{a red \{\}}" is modified to "\prompt{\{\} Make it 
 red}"

\section{Qualitative Results}

\begin{figure}[h]
	\centering
	\includegraphics[width=0.85\linewidth]{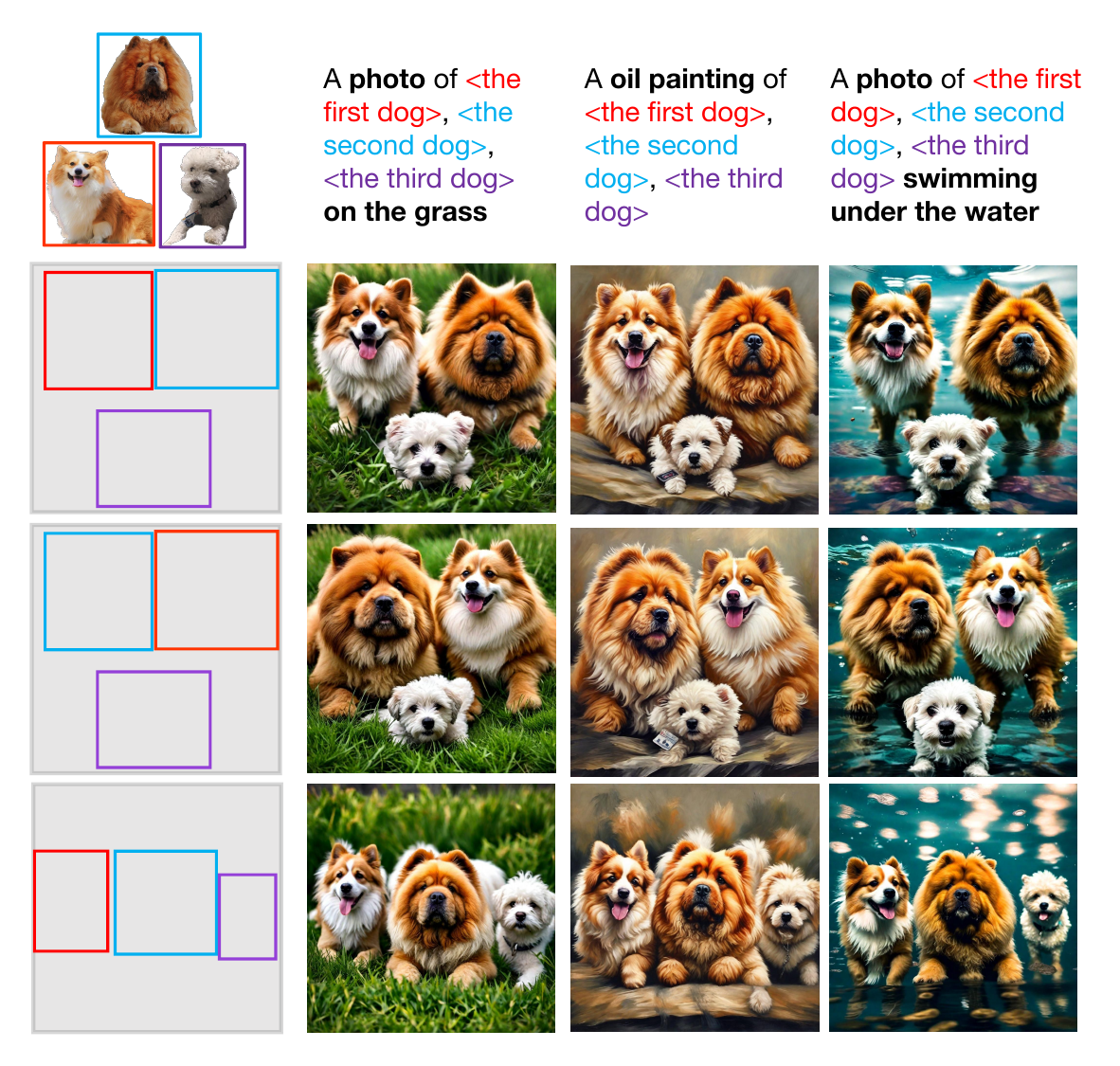}
	\caption{Illustration of controllable visual generation of subject-driven generation across multiple images with layout guidance.}
	\label{fig:method_gen}
\end{figure}

\begin{figure*}[t]
	\centering
	\includegraphics[width=0.999\linewidth]{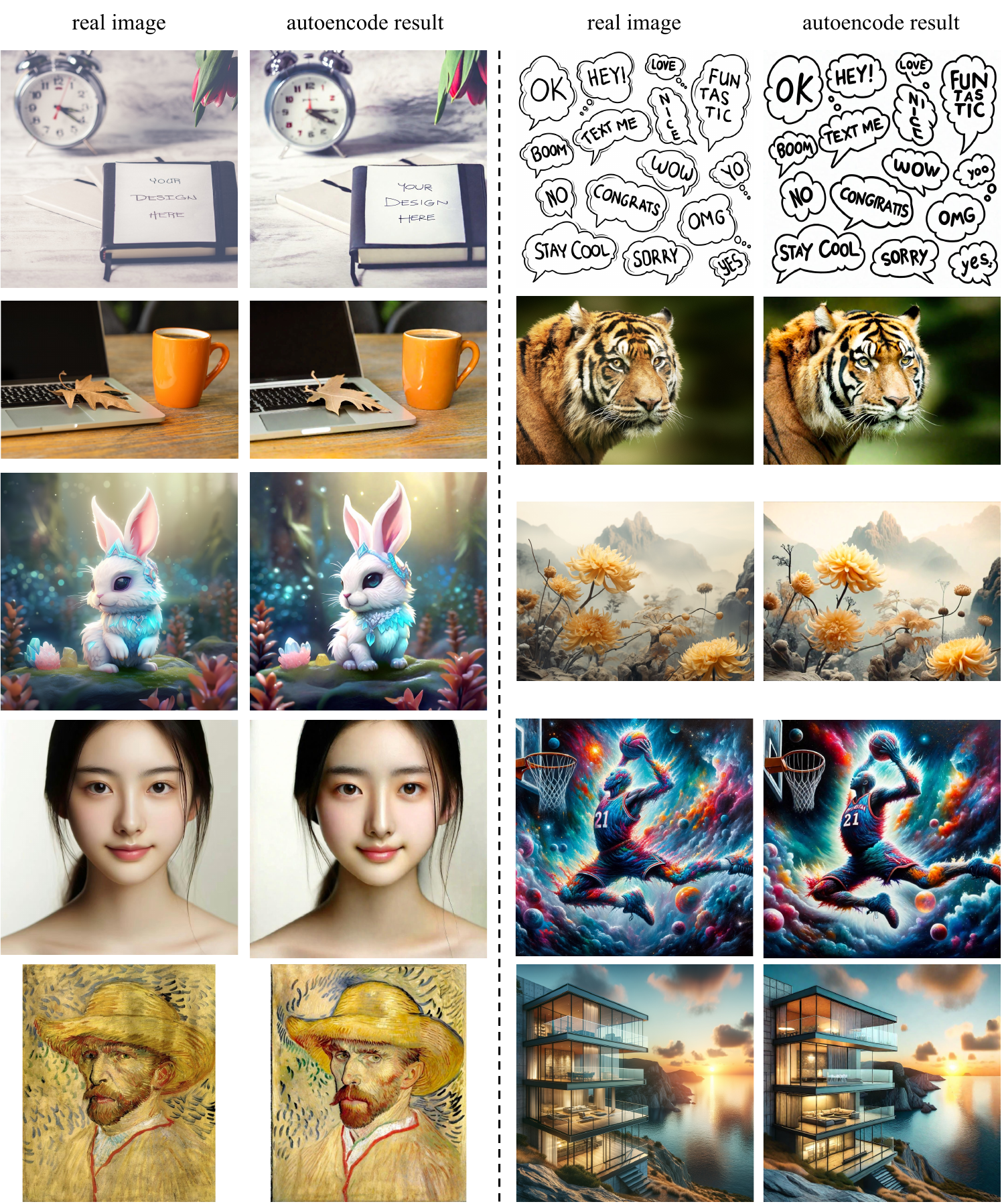}
	\caption{Qualitative cases of image autoencoding.}
\label{fig:supp_ae_cases}
\end{figure*}

\begin{figure*}[t]
	\centering
	\includegraphics[width=0.999\linewidth]{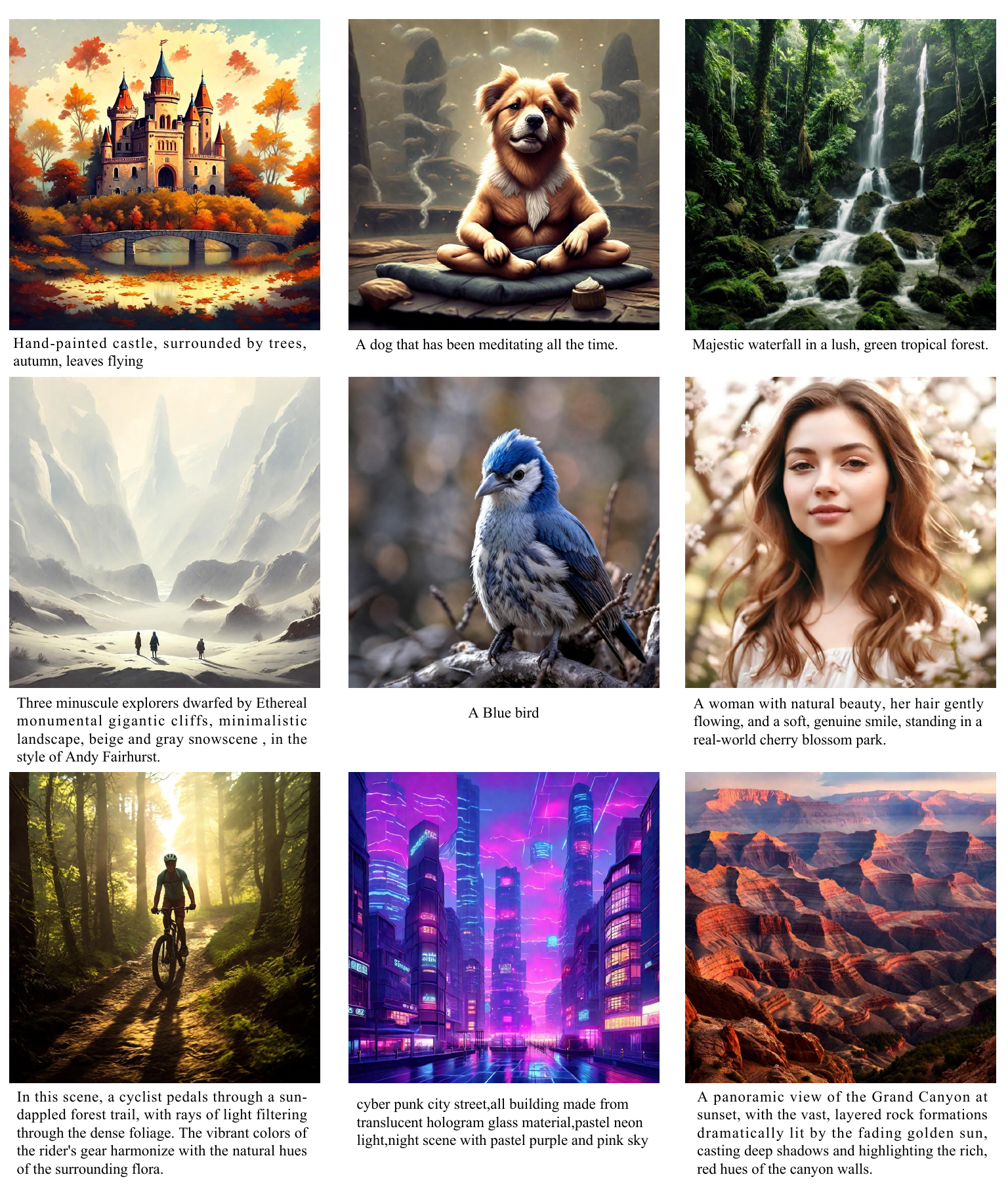}
	\caption{Illustration of text-to-image generation.}
\label{fig:supp_t2i_cases}
\end{figure*}

\begin{figure*}[t]
	\centering
	\includegraphics[width=0.999\linewidth]{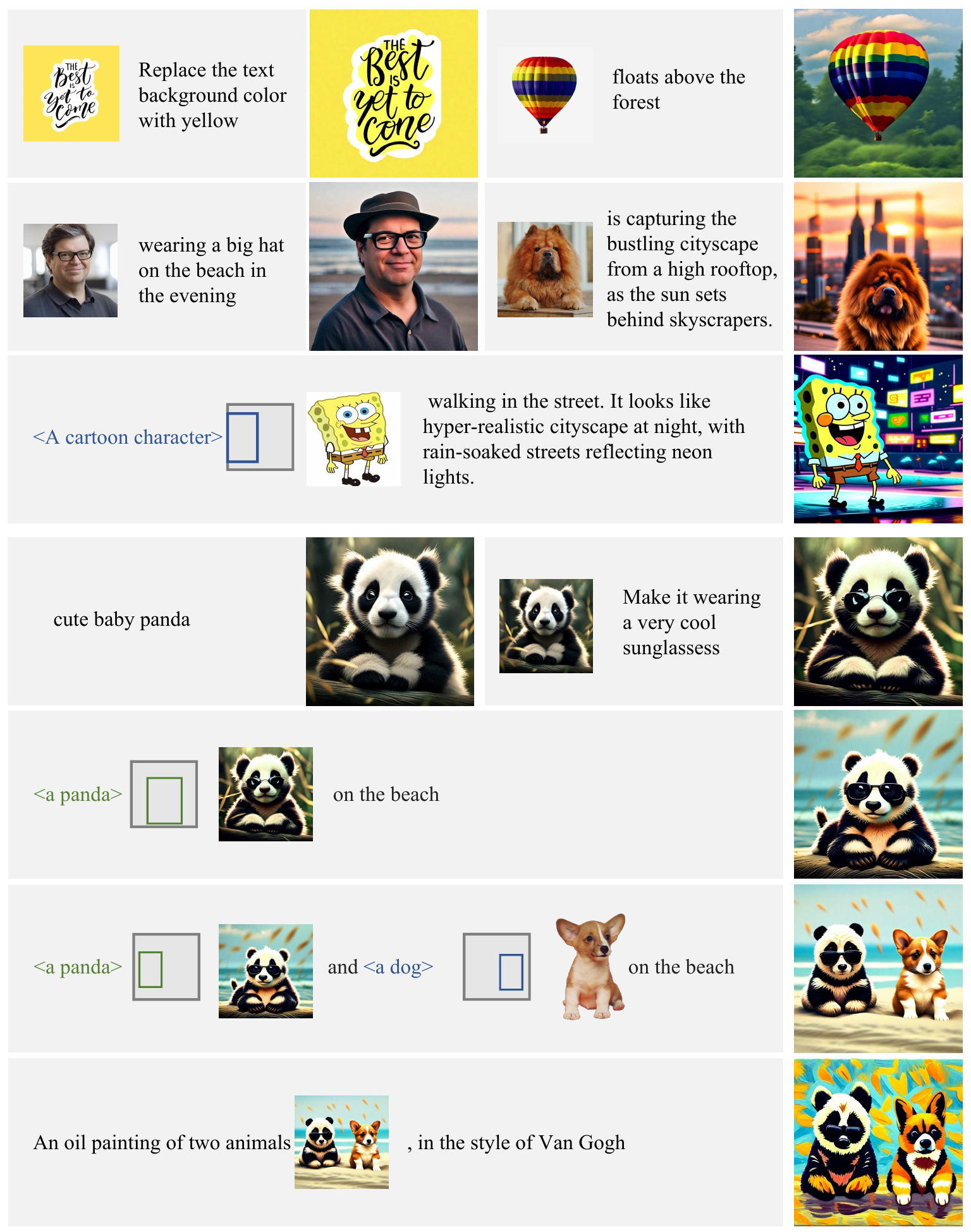}
	\caption{Illustration of zero-shot controllable visual generation with interleaved vision-language prompt.}
	\label{fig:method_gen_edit}
\end{figure*}

\begin{figure*}[t]
	\centering
	\includegraphics[width=0.999\linewidth]{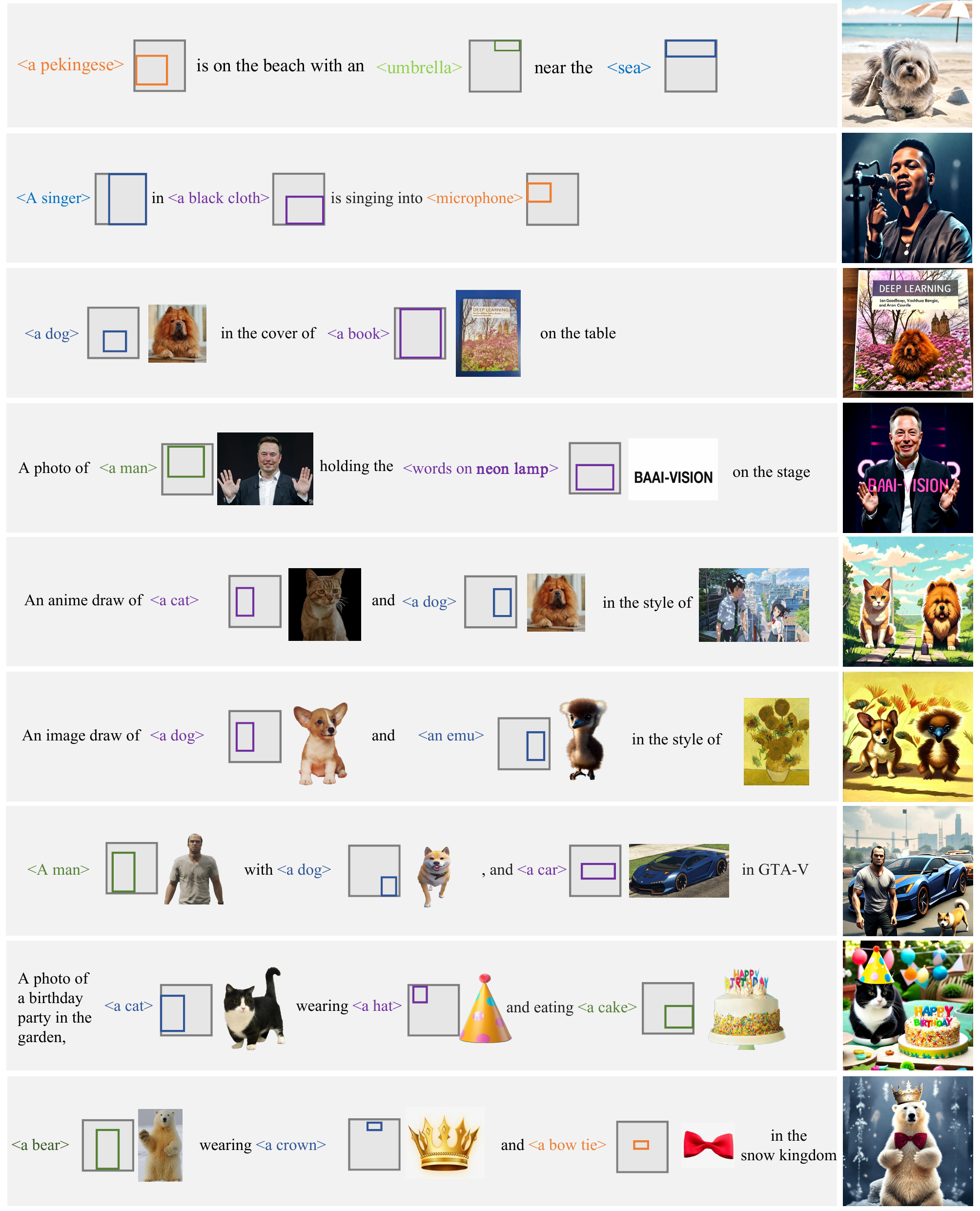}
	\caption{Illustration of zero-shot controllable visual generation with interleaved vision-language prompt.}
	\label{fig:method_gen_v3}
\end{figure*}

\begin{figure*}[t]
	\centering
	\includegraphics[width=0.999\linewidth]{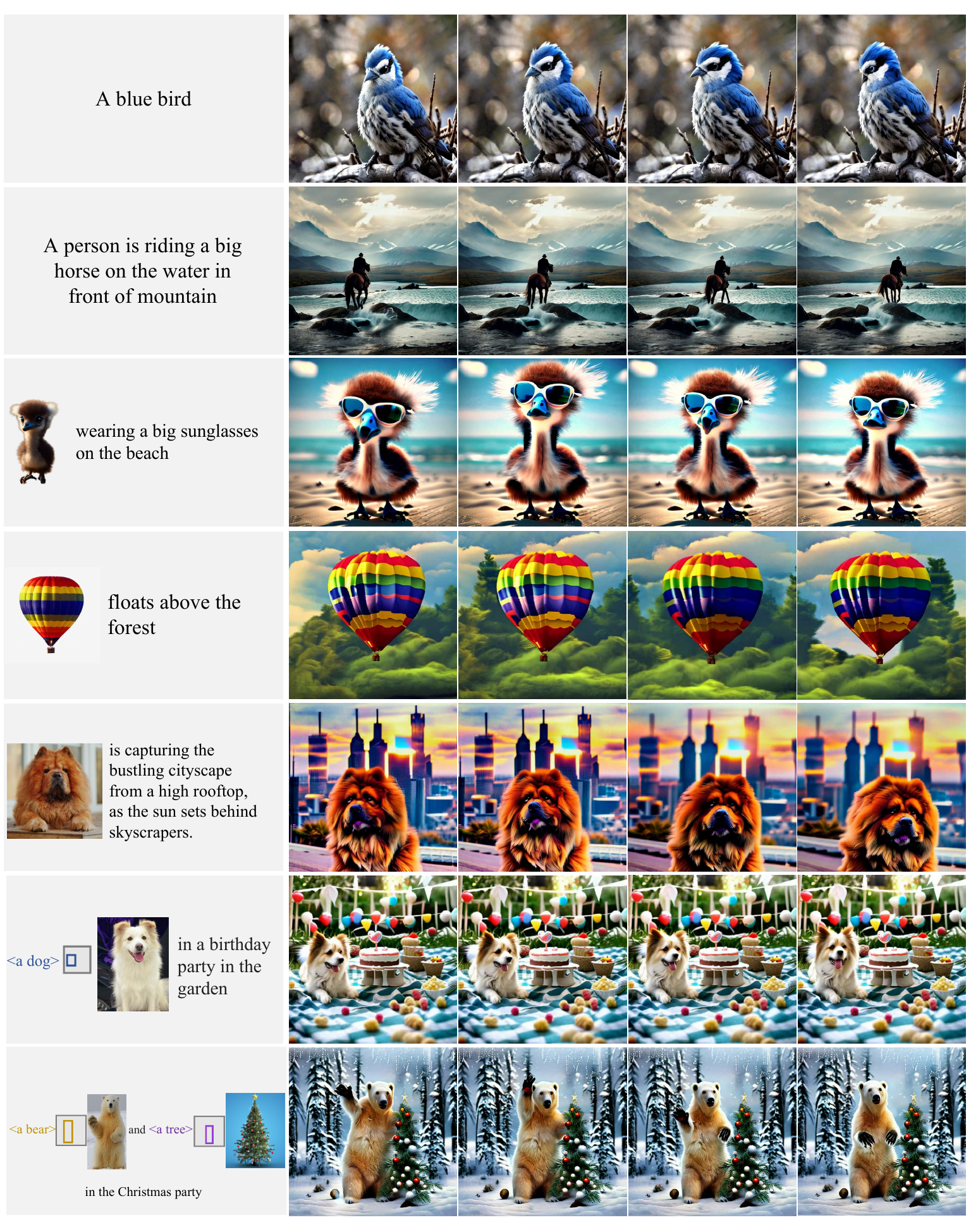}
	\caption{Illustration of zero-shot video generation with interleaved vision-language prompt.}
	\label{fig:method_gen_video}
\end{figure*}

\begin{figure*}[t]
	\centering
	\includegraphics[width=0.9\linewidth]{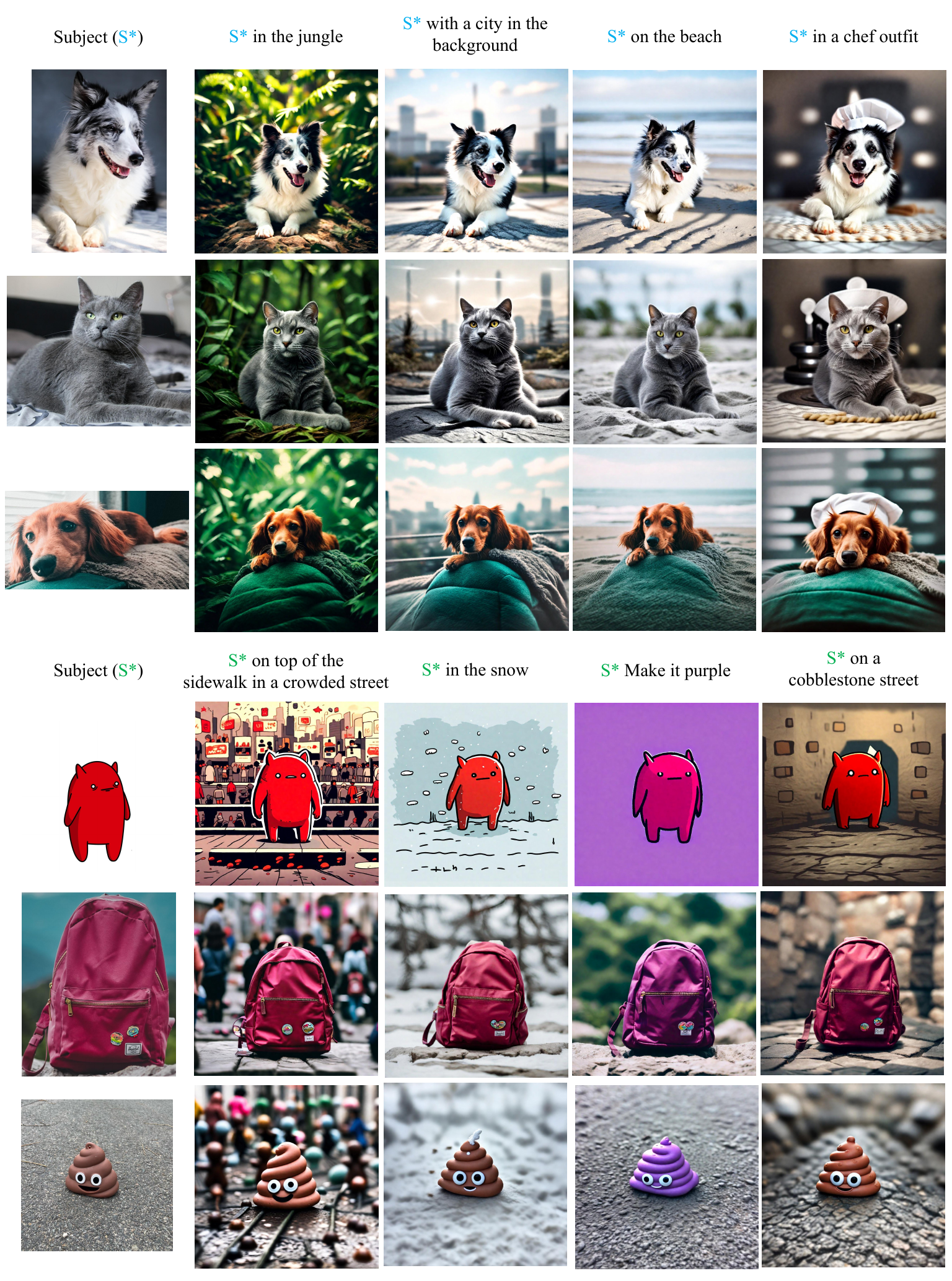}
	\caption{Zero-shot subject-driven generation results on DreamBench.}
\label{fig:supp_dreambooth_cases}
\end{figure*}

\begin{figure*}[!h]
	\centering
	\includegraphics[width=0.999\linewidth]{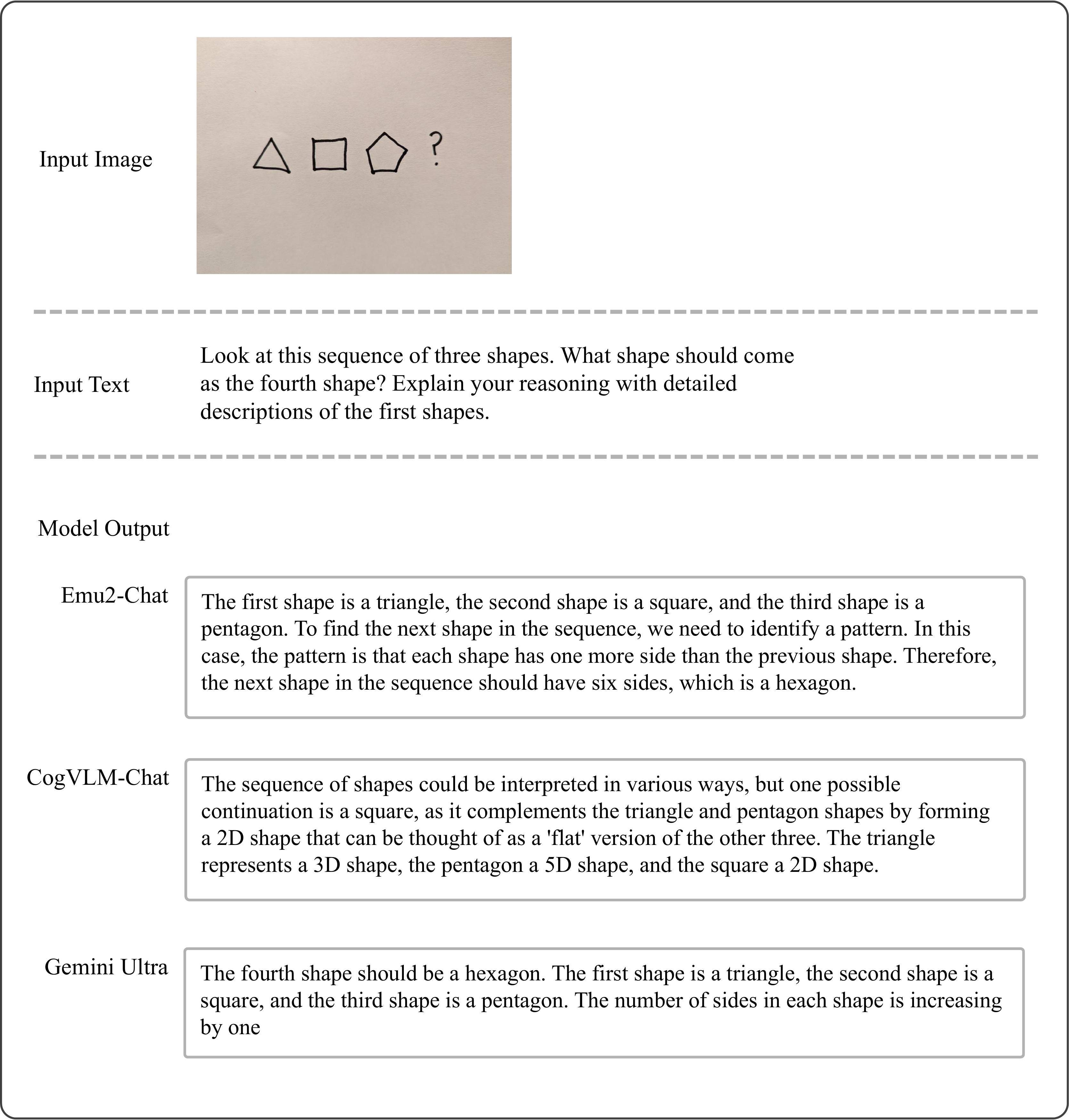}
	\caption{Qualitative case of \OursChat.}
\label{suppfig:chatcase1}
\end{figure*}

\begin{figure*}[!h]
	\centering
	\includegraphics[width=0.999\linewidth]{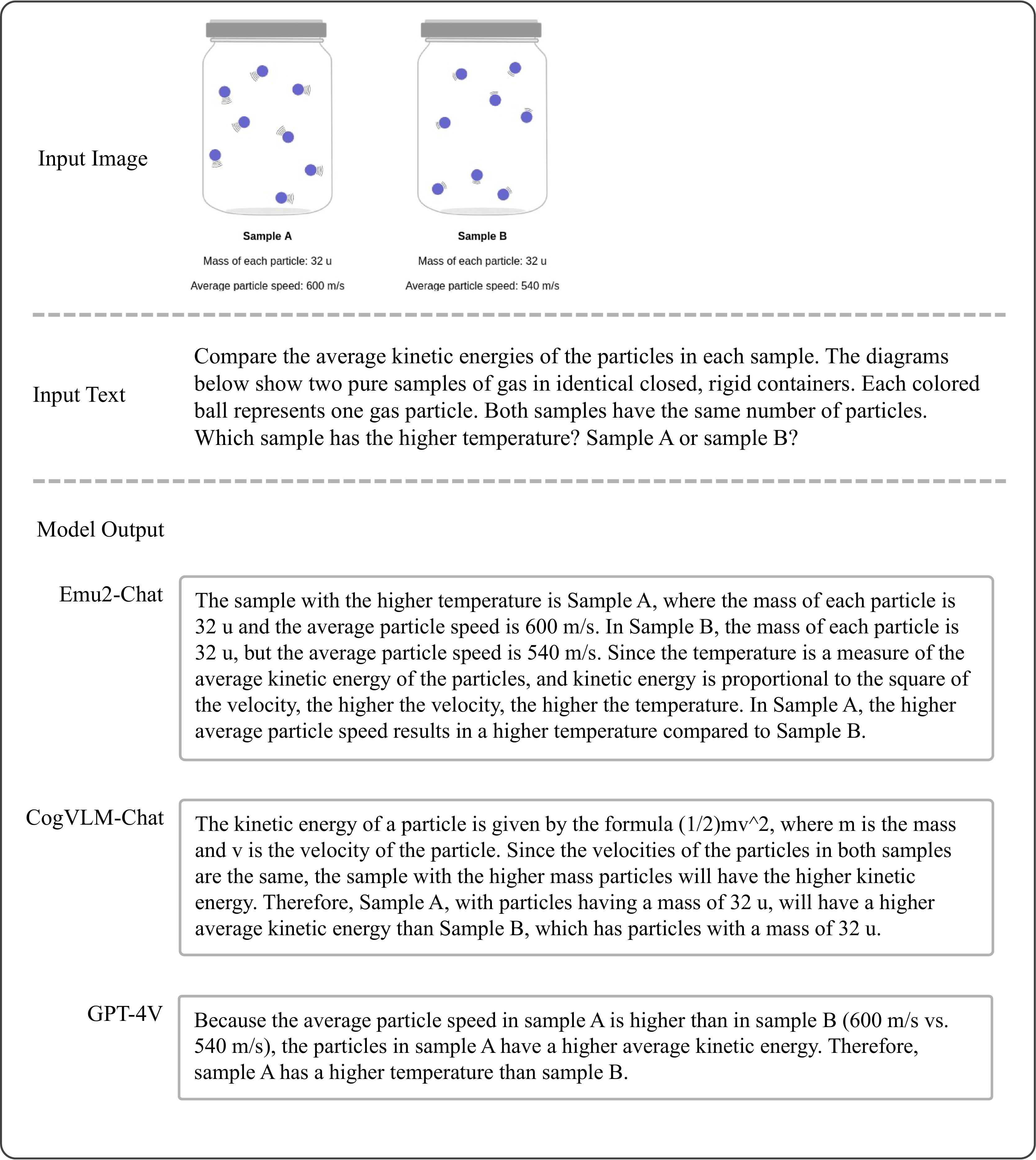}
	\caption{Qualitative case of \OursChat.}
\label{suppfig:chatcase2}
\end{figure*}

\begin{figure*}[!h]
	\centering
	\includegraphics[width=0.999\linewidth]{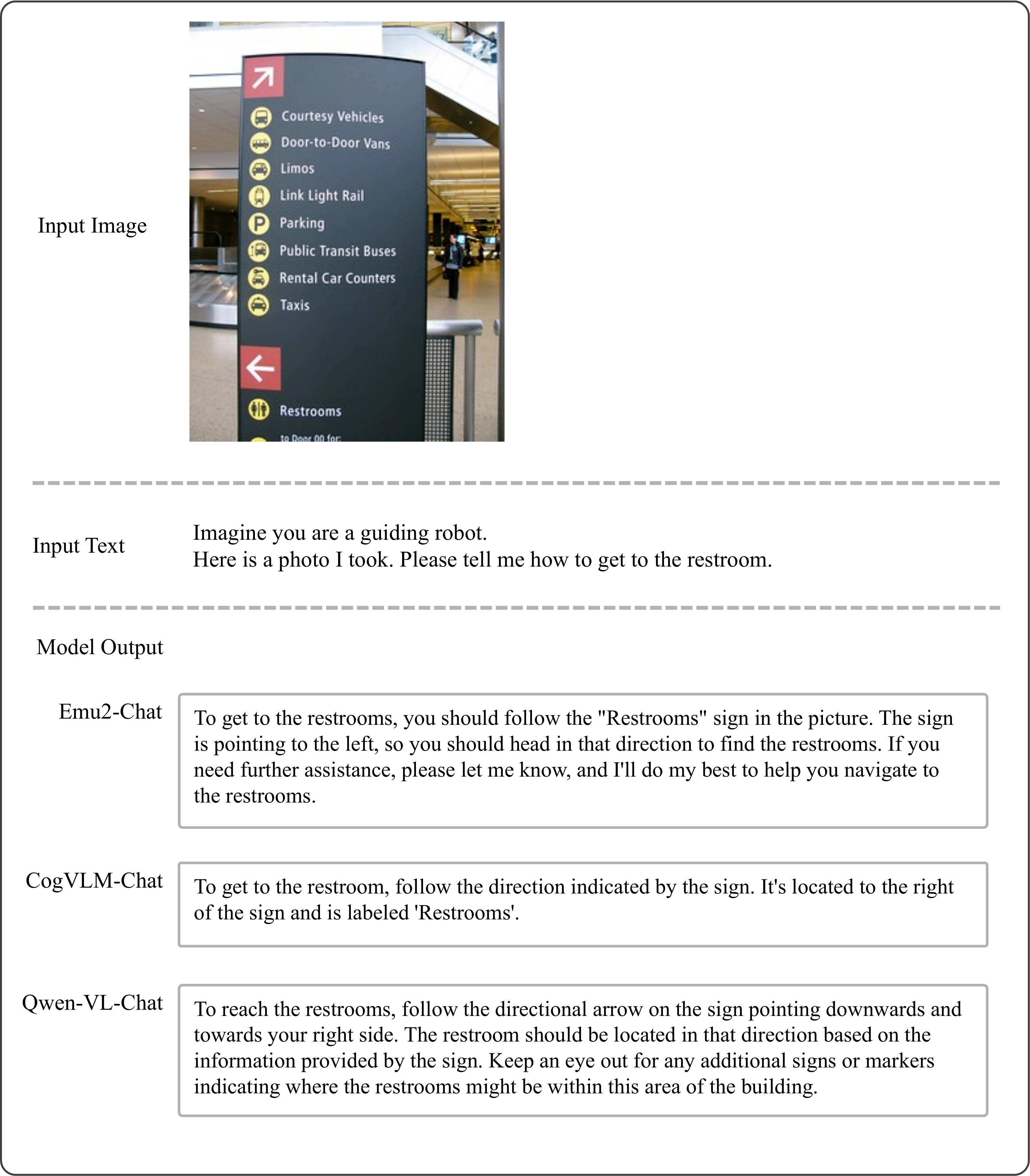}
	\caption{Qualitative case of \OursChat.}
\label{suppfig:chatcase3}
\end{figure*}

We present qualitative cases for \OursG in Figure~\ref{fig:method_gen}-\ref{fig:supp_dreambooth_cases} and for \OursChat in Figure~\ref{suppfig:chatcase1}-\ref{suppfig:chatcase3}, respectively.

\end{document}